%% file: main.tex
\documentclass[acmtog,screen]{acmart}
\acmSubmissionID{431}

\usepackage{booktabs} 
\usepackage{float}
\usepackage{multirow}
\usepackage[ruled]{algorithm2e} 

\SetAlFnt{\small}
\SetAlCapFnt{\small}
\SetAlCapNameFnt{\small}
\SetAlCapHSkip{0pt}

\usepackage{comment}

\setcopyright{acmlicensed}

\acmJournal{TOG}
\acmYear{2024}
\acmVolume{43}
\acmNumber{6}
\acmArticle{}
\acmMonth{12}

\acmDOI{10.1145/3687921}

\newcommand{\methodName}{GarVerseLOD\xspace}
\newcommand{\dataName}{GarVerseLOD\xspace}

\begin{document}

\title{GarVerseLOD: High-Fidelity 3D Garment Reconstruction from a Single In-the-Wild Image using a Dataset with Levels of Details}

\author{Zhongjin Luo}
\orcid{0000-0002-3483-4236}
\affiliation{%
  \institution{SSE, CUHKSZ}
  \city{Shenzhen}
  \country{China}
}

\author{Haolin Liu}
\orcid{0009-0009-4962-9217}
\affiliation{%
  \institution{FNii, CUHKSZ}
  \country{China}
}
\affiliation{%
  \institution{SSE, CUHKSZ}
  \country{China}
}

\author{Chenghong Li}
\orcid{0009-0004-0604-7421}
\affiliation{%
  \institution{FNii, CUHKSZ}
  \country{China}
}
\affiliation{%
  \institution{SSE, CUHKSZ}
  \country{China}
}

\author{Wanghao Du}
\orcid{0009-0005-3005-5007}
\affiliation{%
  \institution{SSE, CUHKSZ}
  \country{China}
}

\author{Zirong Jin}
\orcid{0009-0006-0558-8781}
\affiliation{%
  \institution{SSE, CUHKSZ}
  \country{China}
}

\author{Wanhu Sun}
\orcid{0009-0008-5740-9997}
\affiliation{%
  \institution{SSE, CUHKSZ}
  \country{China}
}

\author{Yinyu Nie}
\orcid{0000-0001-7023-6797}
\affiliation{%
  \institution{Huawei Noah’s Ark Lab}
  \country{UK}
}

\author{Weikai Chen}
\orcid{0000-0002-3212-1072}
\affiliation{%
  \institution{DCC Algorithm Research Center, Tencent Games}
  \country{USA}
}

\author{Xiaoguang Han}
\authornote{Corresponding author is Xiaoguang Han (hanxiaoguang@cuhk.edu.cn).}
\orcid{0000-0003-0162-3296}
\affiliation{%
  \institution{SSE, CUHKSZ}
  \city{Shenzhen}
  \country{China}
}
\affiliation{%
  \institution{FNii, CUHKSZ}
  \city{Shenzhen}
  \country{China}
}

\begin{teaserfigure}
  \centering
  \includegraphics[width=.99\linewidth]{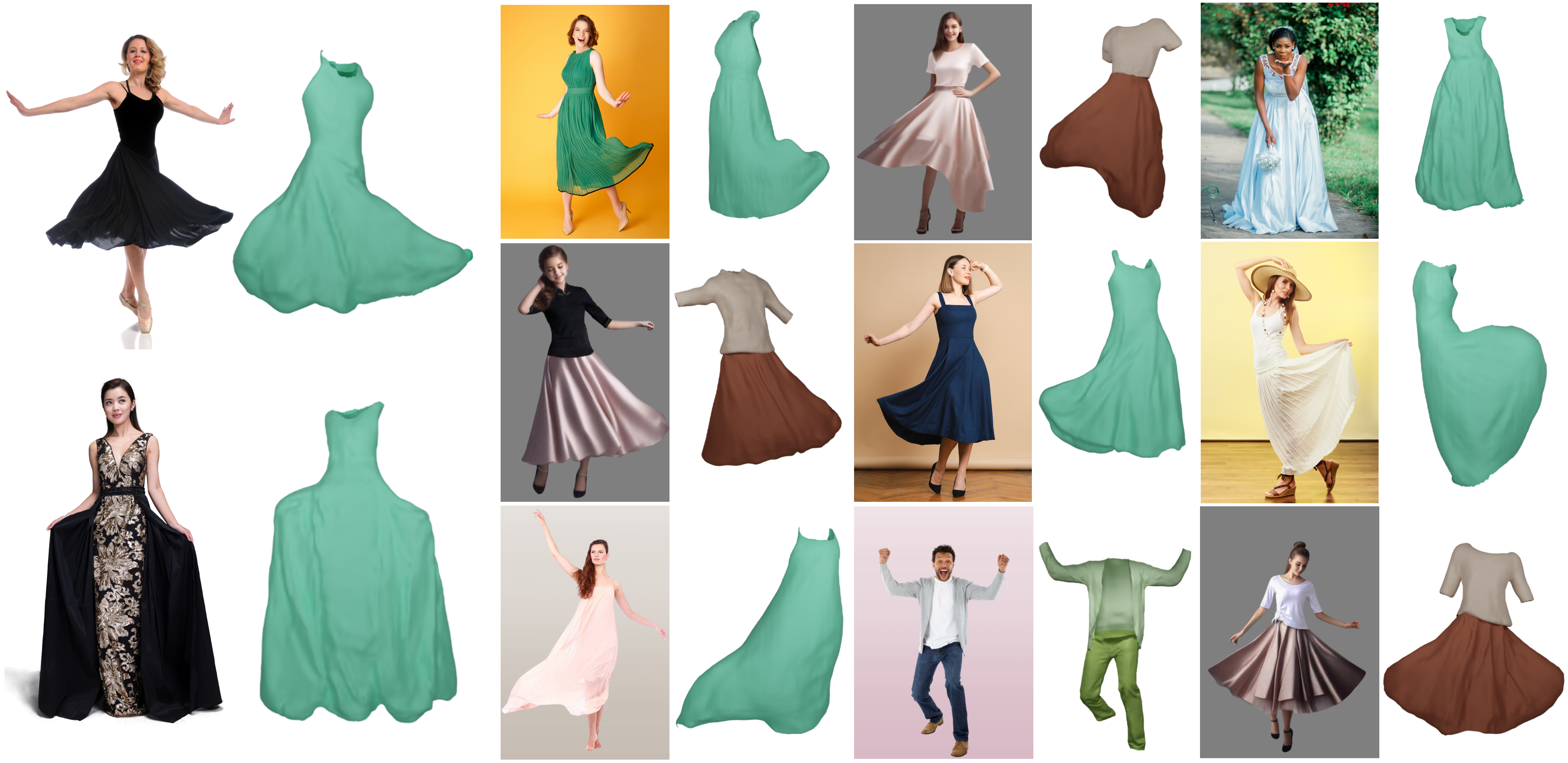}
  \caption{We propose a hierarchical framework to recover different levels of garment details by leveraging the garment shape and deformation priors from the \dataName dataset. Given a single clothed human image, our approach is capable of generating high-fidelity 3D standalone garment meshes that exhibit realistic deformation and are well-aligned with the input image. Original images courtesy of licensed photos and Stable Diffusion~\cite{rombach2022high}. The images with a gray background are synthesized, while the rest are licensed photos.}
  \label{fig_teaser}
\end{teaserfigure}

\input{section/0_abstract}



\maketitle
\input{section/1_intro}
\input{section/2_related}
\input{section/3_dataset}

\input{section/4_method}
\input{section/5_experiment}

\input{section/6_conclusion}
\begin{acks}
The work was supported in part by the Basic Research Project No.~HZQB-KCZYZ-2021067 of Hetao Shenzhen-HK S\&T Cooperation Zone, by Guangdong Provincial Outstanding Youth Fund (No.~2023B1515020055), by Shenzhen Science and Technology Program No.~JCYJ20220530143604010, and No.~NSFC\-61931024. 
It is also partly supported by the National Key R\&D Program of China with grant No.~2018YFB1800800, by Shenzhen Outstanding Talents Training Fund 202002, by Guangdong Research Projects No.~2017ZT07X152 and No.~2019CX01X104, by Key Area R\&D Program of Guangdong Province (Grant No.~2018B030338001), by the Guangdong Provincial Key Laboratory of Future Networks of Intelligence (Grant No.~2022B1212010001), and by Shenzhen Key Laboratory of Big Data and Artificial Intelligence (Grant No.~ZDSYS201707251409055). Stable Diffusion~\cite{rombach2022high} was utilized to generate the synthetic images used in this work.
\end{acks}

\input{figures/fig_result_gallery}
\input{figures/fig_compare}

\clearpage
\input{figures/fig_ablation_boundary}

\input{figures/fig_alation_data}

\input{figures/fig_ablation_coarse}

\input{figures/fig_ablation_implicit_udf}

\input{figures/fig_limitation}

\clearpage

\bibliographystyle{ACM-Reference-Format}
\bibliography{reference}

\clearpage

\appendix
\input{supp/8_implentation}
\input{supp/7_data}

\clearpage
\input{supp/9_more_data_more_results}

\clearpage
\input{supp_tables/table_notation}

\end{document}

%% file: section/0_abstract.tex
\begin{abstract}
	
Neural implicit functions have brought impressive advances to the state-of-the-art of clothed human digitization from multiple or even single images. However, despite the progress, current arts still have difficulty generalizing to unseen images with complex cloth deformation and body poses. In this work, we present \methodName{}, a new dataset and framework that paves the way to achieving unprecedented robustness in high-fidelity 3D garment reconstruction from a single unconstrained image. Inspired by the recent success of large generative models, we believe that one key to addressing the generalization challenge lies in the quantity and quality of 3D garment data. Towards this end, \methodName{} collects 6,000 high-quality cloth models with fine-grained geometry details manually created by professional artists. In addition to the scale of training data, we observe that having disentangled granularities of geometry can play an important role in boosting the generalization capability and inference accuracy of the learned model.	We hence craft \methodName{} as a hierarchical dataset with \emph{levels of details (LOD)}, spanning from detail-free stylized shape to pose-blended garment with pixel-aligned details. This allows us to make this highly under-constrained problem tractable by factorizing the inference into easier tasks, each narrowed down with smaller searching space. To ensure \methodName{} can generalize well to in-the-wild images, we propose a novel labeling paradigm based on conditional diffusion models to generate extensive paired images for each garment model with high photorealism. 
We evaluate our method on a massive amount of in-the-wild images. Experimental results demonstrate that \methodName{} can generate standalone garment pieces with significantly better quality than prior approaches while being robust against a large variation of pose, illumination, occlusion, and deformation. Code and dataset are available at \href{https://garverselod.github.io/}{\textcolor{blue}{garverselod.github.io}}.

\end{abstract}

%% file: section/1_intro.tex
\section{Introduction}

\label{sec:intro}

High-quality 3D garment models are critical assets for a large variety of applications, ranging from entertainment to professional concerns, such as visual effects, physical simulation, and VR/AR telepresence.
In the production-level pipeline, independent garment pieces are more desirable than a single clothed human model, as the former allows layered compositions with an internal body mesh to ensure the realism of physical motion and the flexibility of garment transfer.
However, unlike clothed human reconstruction that can directly utilize the latest advances of neural implicit representation~\cite{saito2019pifu,saito2020pifuhd,xiu2022icon}, standalone garment modeling mostly relies on deforming parametric templates with open boundaries due to its strict requirement of correct topology.


Nonetheless, reconstructing high-fidelity 3D garment from a single image remains a nuisance to current vision algorithms.
While the high diversity of garment styles and the scarcity of the inputs render the problem highly ill-posed, the complex deformations resulted from the cloth dynamics make the inference even more challenging.
There are two mainstream approaches for estimating the deformations of standalone garments from posed humans.
Linear blend skinning (LBS)-based methods~\cite{jiang2020bcnet,corona2021smplicit} focus on predicting the deformations caused by human poses, where the learned skinning weights of the garment mesh are either bound to the skeleton or the surface vertices of a parametric model of unclothed humans (e.g., SMPL~\cite{loper2015smpl}). 
While this line of approaches can effectively represent posed-induced deformations, they struggle to model other intricate deformations caused by the environments or physical dynamics.
Feature-line-based methods~\cite{zhu2020deep,zhu2022registering} reconstruct garment meshes from SMPL surfaces and further fit them with garments' manifold boundaries, making it versatile to model any type of deformations. However, the problem of boundary estimation from single images itself is challenging, due to the severe occlusions and 2D-to-3D ambiguities.

Apart from the technical challenges, the other obstacle to learning-based garment reconstruction is the limited quantity and quality of 3D dataset. 
Due to the lack of local geometry details in existing garment datasets, current LBS-based methods are incapable of learning fine-grained geometries (e.g., wrinkles), resulting in coarse 3D garment quality. 
ReEF~\cite{zhu2022registering} annotates the feature lines for only 400 garment models in the RenderPeople dataset~\cite{RenderPeople}. The limited data scale hampers the prior approaches from generalizing to unseen images and often leads to poor reconstruction quality of feature lines (i.e., garment boundaries).

In this work, we strive to address the above issues for standalone 3D garment reconstruction from the perspectives of both data and algorithm.
We thereby introduce \methodName{}, a dedicated dataset and framework that achieves unprecedented robustness in reconstructing high-fidelity 3D garments from a single in-the-wild image (Fig.~\ref{fig_teaser}). 
To promote the quantity and quality of 3D garment data, \methodName{} collects 6,000 high-quality hand-crafted garment meshes with fine-grained details created by professional artists. 
It covers 5 most commonly seen categories -- each category shares the same mesh topology, facilitating cross-instance interpolation and construction of blendshape models.
While garment shapes differ globally in terms of style and topology, the local deformations are determined by a wide range of factors, including body poses, garment-environment interactions, self-collisions, \emph{etc}. 
We, therefore, propose to craft \methodName{} as a hierarchical dataset with \emph{levels of details (LOD)} to accommodate this key observation.

In particular, as shown in Fig.~\ref{fig:data_pipeline}, \methodName{} contains three basic levels of databases: 1) \emph{Garment Style Database} with T-posed and detail-free coarse garment; 2) \emph{Local Detail Database} enclosing pairs of T-posed models with and without fine-level local geometric details; and 3) \emph{Garment Deformation Database} consisting of pairs of T-posed garment and its deformed counterpart (i.e., with global deformations).
As the mesh topologies are identical within each category, we can easily extract the local details and global deformations from paired models in the corresponding database and combine all levels of geometries to obtain the \emph{Fine Garment Dataset}. 
The disentangled granularities of geometry allows us to make this highly underconstrained problem tractable by factorizing the inference into smaller tasks, each can be tackled with narrowed solution space.  
Furthermore, we introduce a novel data labeling paradigm to generate extensive paired images for each garment model. 
Specifically, we leverage the latest advances in conditional diffusion model to transfer the textureless renderings to photorealistic images with diverse appearances. This further elevates the generalization capability of \methodName{} in handling unconstrained images.



Algorithm-wise, we propose to connect the good ends of both LBS and feature-line based approaches. 
We first build a parametric model of the T-posed coarse shapes in the garment style database. 
After estimating the blendshape coefficients of the coarse garment, we progressively refine the result by adding pose-induced global deformations and fine-scale local deformations. 
Thanks to the LOD structure of \methodName{}, these three steps can be performed in a disentangled manner with eased complexity. 
While we employ linear blend skinning to estimate deformations caused by body poses, an implicit garment representation is learned to capture pixel-aligned fine surface from estimated 2D normal maps.
We then fit the posed coarse garment with fine surfaces by aligning their open boundaries for the purpose of transferring the local details to the globally deformed mesh with correct topology.
To combat with the occlusions, we present a novel geometry-aware boundary prediction strategy that equips the 2D features with 3D information from the estimated fine surface for better localization of 3D boundaries.
Our experimental results show that \methodName
can effectively reconstruct garments with diversified shapes and intricate deformations, demonstrating significantly better generalization ability over the prior arts. We summarize our contributions as follows:

\begin{itemize}
\item We present the \methodName dataset, a large collection of high-fidelity 3D \emph{hand-crafted} garments. It encloses 6,000 professionally hand-crafted garments, covers 5 categories, and, for the first time, contains 3 disentangled levels of details to ease the learning task.


\item We propose a novel data simulation pipeline to generate extensive paired images for supporting single-view reconstruction.

\item We devise a specially-tailored coarse-to-fine approach to fully utilize the LOD structure of the \methodName{} dataset. Experimental results show that our method excels in reconstructing high-quality garments from single images.

\end{itemize}

%% file: section/2_related.tex
\section{Related Work}
\label{sec:related}

\paragraph{3D Human Reconstruction.}

3D reconstruction has seen significant advancements recently~\cite{loper2015smpl,saito2019pifu,luo2021simpmodeling,poole2022dreamfusion,yan2024dreamdissector,luo2023rabit,habermann2019livecap,habermann2020deepcap,li2021deep,jiang2022hifecap,xu2018monoperfcap}.
Some single-view human reconstruction methods~\cite{lassner2017unite,pavlakos2019expressive,xu2019denserac,anguelov2005scape,hasler2009statistical} restrict the solution space to a parametric human model and simplify the problem, which can only reconstruct nude human 3D models without garments.
Inspired by SMPL~\cite{loper2015smpl}, some methods~\cite{alldieck2019tex2shape,tan2020self,xiang2020monoclothcap,yang2018physics,zheng2019deephuman} approximate human body geometry by deforming the SMPL. These methods can reconstruct realistic results from an unconstrained image but fail to handle loose garments.
Contrary to the SMPL-based approaches, other methods enable clothed human body reconstruction with arbitrary topology.
Siclope~\cite{natsume2019siclope} reconstructs clothed 3D human models using multi-view silhouettes predicted from a frontal image.
DeepHuman~\cite{zheng2019deephuman} generates progressively refined voxels, which are embossed with details from a surface normal. 
While both methods can produce clothed human shapes with arbitrary topology, the details are relatively coarse.
Recent works~\cite{li2020monocular,saito2019pifu,saito2020pifuhd,xiu2022icon,xiu2023econ} address this issue with pixel-aligned implicit functions and achieve high reconstruction fidelity. 
However, all the above methods fail to provide the garment mesh separated from the human body.

\paragraph{3D Garment Reconstruction.} 
Compared to clothed 3D human bodies, reconstructing high-fidelity 3D independent garments from a single image is challenging. Many prior arts rely on learning-based strategies to fit 3D garment deformations from a collection of 2D image-3D garment pairs for generalization. Two mainstream approaches are often used in estimating standalone garments from single images. 
Linear blend skinning (LBS)-based methods (e.g., BCNet~\cite{jiang2020bcnet}, ClothWild~\cite{moon20223d}) focus on predicting deformations caused by human poses, where explicit or implicit garment parametric models are used. These methods can address posed-guided deformations but fail to reproduce large cloth deformations (e.g., those caused by complex environmental factors) and fine-grained surface details (e.g., wrinkles). 
Recent works, such as ISP~\cite{li2024isp} and Neural-ABC~\cite{chen2024neural}, have proposed more advanced implicit parametric models, but their reconstruction methods still rely solely on their parametric models. Similar to BCNet~\cite{jiang2020bcnet} and ClothWild~\cite{moon20223d}, the limited representational capacity of the parametric models prevents them from accurately recovering complex garment deformations from images. 
Feature-line-based methods~\cite{zhu2020deep,zhu2022registering} reconstruct garment meshes from SMPL and further fit them with the estimated pixel-aligned clothed human and the predicted garment's manifold boundaries, making them flexible to model cloth deformations and geometric details. However, the problem of boundary estimation from single images itself is challenging, due to the severe occlusions and 2D-to-3D ambiguities. 
Garment Recovery~\cite{li2024garment} relies on a normal estimator trained on human data with limited clothing diversity and a deformation prior trained with limited deformation variations, preventing it from accurately reconstructing high-fidelity surface details and complex deformations that reflect the inputs.
All existing methods cannot faithfully recover intricate clothing deformations and fine-grained geometric details from single-view images.


\input{figures/fig_dataset_pipeline}

\paragraph{3D Garment Datasets.}
3D garment datasets are an important foundation for learning-based tasks, but current available datasets are very limited in quality and scale. Existing datasets can be divided into two major categories: scanning-based datasets~\cite{zhang2017detailed,pons2017clothcap,bhatnagar2019multi,zhu2020deep, tiwari2020sizer,lin2023leveraging,wang20244d} and simulation-based datasets~\cite{patel20tailornet,jiang2020bcnet,gundogdu2019garnet,bertiche2020cloth3d,zou2023cloth4d,black2023bedlam}.
Scanning-based datasets allow for realistic garment appearance and shape. However, separating garment models from 3D scans is laborious and often results in surface damage due to occlusion. These datasets usually are restricted by data scale and suffer from the inability to separate garments from the mannequin~\cite{zhang2017detailed}, as well as insufficient clothing diversity. Simulation-based datasets synthesize 3D garments by simulating motion using physics-based engines. However, these synthetic datasets are unsatisfactory in terms of cloth style, body pose and garment deformation variations, as well as the quality of paired images. It is difficult to generalize the trained models to in-the-wild images. We introduce a large-scale 3D garment dataset characterized by intricate deformations and fine-grained surface details. Additionally, we present a novel data simulation strategy to collect extensive image-3D garment pairs by leveraging the generative capabilities of conditional stable diffusion models~\cite{rombach2022high,mou2023t2i,zhang2023adding}.

%% file: figures/fig_dataset_pipeline.tex
\begin{figure*}[htbp]
  \centering
  \includegraphics[width=.96\linewidth]{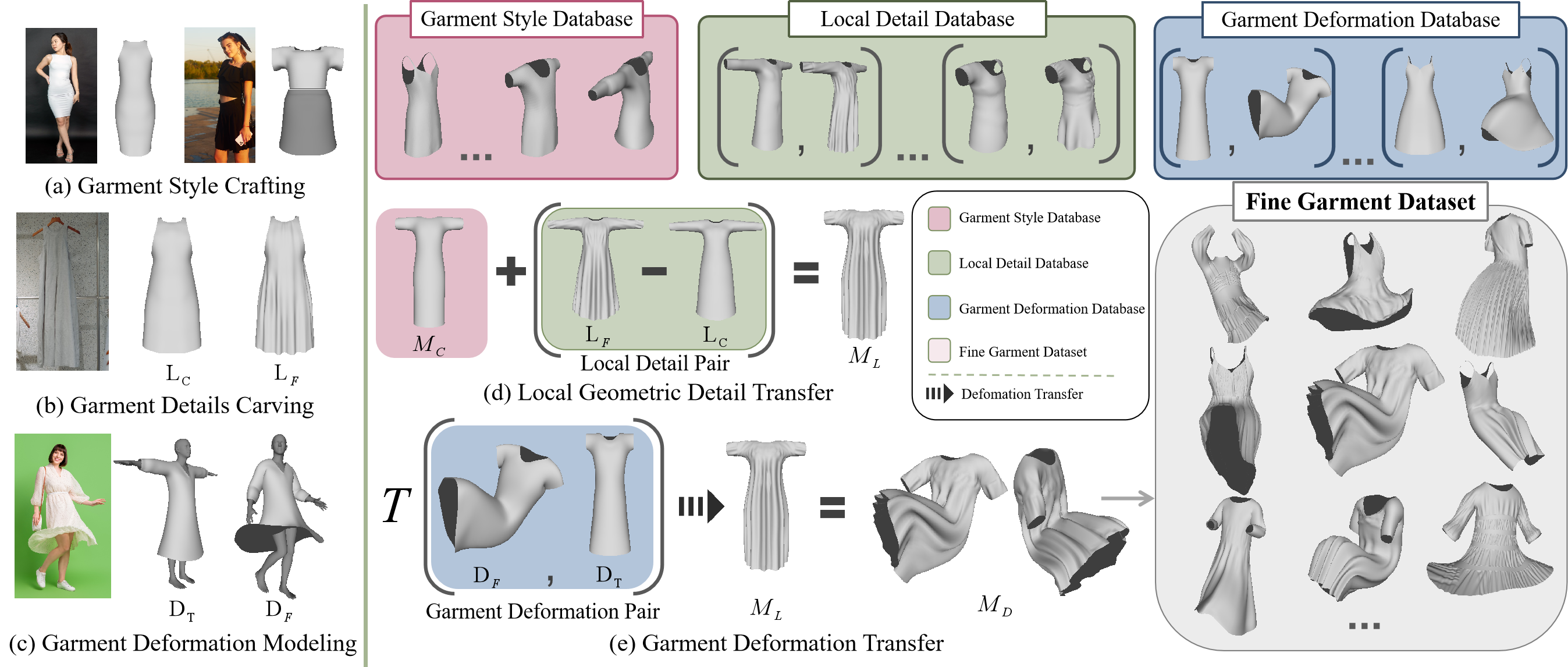}
  \caption{The pipeline of our novel strategy for constructing a progressive garment dataset with levels of details. (a) Each case shows the reference image and the artist-crafted T-pose coarse garment in \textbf{Garment Style Database}. (b) A example of the reference image and the artist-crafted detail-pair in \textbf{Local Detail Database}. (c) A example of the reference image and the artist-crafted deformation-pair in \textbf{Garment Deformation Database}. (d) To obtain an T-pose garment with geometric details, we first sample a shape $M_C$ from the Garment Style Database and a ``Local Detail Pair'' ($L_C$, $L_F$) from the Local Detail Database. Then we transfer the geometric details depicted by ($L_C$, $L_F$) to $M_C$ to obtain $M_L$. (e) The deformation depicted by a sampled ``Garment Deformation Pair'' ($D_T$, $D_F$) is transferred to $M_L$ to obtain the fine garment $M_D$, which contains fine-grained geometric details and complex deformations (\textbf{Fine Garment Dataset}). Original images courtesy of licensed photos.
  }
  \label{fig:data_pipeline}
\end{figure*}

%% file: section/3_dataset.tex
\section{Dataset}
\label{sec:dataset}
Reconstructing accurate and standalone garments from single images remains a significant challenge due to the absence of a well-established dataset, especially for scenarios involving complex deformations like human-garment or garment-environment interactions. Existing datasets suffer from limited data scale~\cite{jiang2020bcnet,bhatnagar2019multi}, or a lack of paired 2D image-3D garment data~\cite{bertiche2020cloth3d}, or solely monotonous rendered images paired with clothing models~\cite{zou2023cloth4d}. In this work, we fill this gap by introducing \dataName,  a progressive dataset with levels of details (LOD). Additionally, we present a novel data generation pipeline to construct a large-scale dataset with realistic images paired with 3D models in \dataName.
\paragraph{Overview.} 
\dataName has four features:
1) \textbf{Broad diversity}. 
\dataName contains 5 common garment categories, i.e. dress, skirt, coat, top, and pant. Each 3D model comprises fine-grained geometric details and intricate clothing physical deformations.
2) \textbf{Levels of details}. 
We collect three basic databases with different \emph{levels of details} (LOD) to obtain high-quality 3D clothes. Based on these databases, we create a large number of posed 3D garments with complex deformations and fine-grained geometric details.
3) \textbf{Topological consistency}. 
Each 3D garment in \dataName is created by carefully deforming a pre-defined template mesh. 
All 3D garments within different categories share a unified topology, paving the way to learn a parametric model. 
4) \textbf{Extensive paired data}. 
To create high-quality image-3D garment paired data, we employ ControlNet~\cite{zhang2023adding} and T2I-Adapter~\cite{mou2023t2i} as the data simulator and transform monotonous rendered images into photorealistic images with diverse appearances. Some 2D image-3D garment pairs are shown in Fig.~\ref{fig:data_gallery}. Please refer to the supplementary materials for more details.

\input{figures/fig_dataset_gallery}

\subsection{LOD Garment Crafting} 
As shown in Fig.~\ref{fig:data_pipeline}, we first construct three basic databases with different levels of detail: 
1) \textbf{Garment Style Database}. In Fig.~\ref{fig:data_pipeline}(a), we collected a set of reference images of clothed humans with diverse garment styles from the Internet and hired eight artists to craft a T-posed coarse garment for each reference image (without surface geometric details like wrinkles, only depicting the overall cloth shape);
2) \textbf{Local Detail Database}. In Fig.~\ref{fig:data_pipeline}(b), we collected a set of reference images of clothes with diverse surface details (e.g., wrinkles). The eight artists were asked to carve two T-posed garments for each image: one without surface details ($L_C$) and one with fine surface details ($L_F$). These garment pairs ($L_C$, $L_F$) describe garment local geometric details;
3) \textbf{Garment Deformation Database}. In Fig.~\ref{fig:data_pipeline}(c), we collected a set of reference images of clothed humans with diverse poses and garment deformations. For each image,  we use PyMAF~\cite{pymaf2021} to estimate the SMPL shape $\beta$ and pose $\theta$ from the images. The artists were asked to construct two over-smoothed garments (i.e., garments without local geometric details): a T-posed garment (i.e, $D_T$ on top of the estimated T-pose body) and a garment with global deformation aligned with the image (i.e., $D_F$, on top of the posed body). These two garments ($D_T$, $D_F$) form a pair that depicts garment deformations. Note that the estimated SMPL parameters are also stored to assist deformation transfer in the following fine garment synthesis. 

\paragraph{\textbf{Fine Garment Dataset}} All models in the above three databases are created by deforming predefined templates (i.e., dress, skirt, coat, top, and pant). Thus, all 3D garments within different categories are homeomorphic in topology. The feature of \textbf{topological-consistency} not only paves the way to learning a parametric model (Sec.~\ref{sec:method_ce}), but also enables the incorporation of our three basic databases to obtain the fine garment dataset. As shown in Fig.~\ref{fig:data_pipeline}, we first sample a coarse garment shape $M_C$ by interpolating between garments in the Garment Style Database. Then we sample a ``Local Detail Pair'' ($L_C$, $L_F$) and apply their vertex offsets to $M_C$ by,
\begin{equation}
M_L = M_C + L_F - L_C,
\end{equation}
where $M_L$ denotes the T-posed garment with local details transferred from $L_F$.
Subsequently, we sample a ``Garment Deformation Pair'' ($D_T$, $D_F$) from the Garment Deformation Database and transfer the deformation to $M_L$ to obtain the fine garment $M_D$ by,
\begin{equation}
M_D = LBS(M_L + T),
\end{equation}
\begin{equation}
T = LBS^{-1}(D_F) - D_T,
\end{equation}
where we apply the inverse LBS of SMPL to garment $D_F$ to obtain the deformation offsets $T$ in the rest-pose space. Then $T$ is applied to $M_L$ in the rest-pose space. The forward LBS is used to deform $M_L$ to pose space to obtain the fine garment $M_D$, which contains both fine-grained surface details and complex garment deformations.

\subsection{Photorealistic Paired Image Generation}

We present a data simulation pipeline to synthesize images paired with our 3D garments. Specifically, we utilize ControlNet and T2I-Adapter to transfer monotonous rendered images to photorealistic images with diverse appearances. As illustrated in Fig.~\ref{fig:data_gallery}, we first obtain 3D garment renderings with random camera views. Then, the rendered images are fed to Canny-Conditional Stable Diffusion (i.e., ControlNet and T2I-Adapter) to obtain realistic RGB images. We calculate the $L_1$ loss on canny edges between renderings and the generated images, and manually pick generated images that closely approximate the 3D garment shape. 
Finally, all generated images are manually inspected to ensure high consistency between the image and the corresponding 3D garment.
\paragraph{Local Alignment} Although ControlNet and T2I-Adapter perform well in generating images with correct global shapes, it is still challenging to produce images with pixel-aligned details, such as wrinkles (see Fig.~\ref{fig:data_gallery}(a, b, d)). To support fine-grained inference, as shown in Fig.~\ref{fig:data_gallery}(c), a pixel-level alignment mask is labeled manually to mark out the alignment region between the synthesized image (b) and the rendered normal map (d). This leads to a collection of high-quality paired data that can be utilized for 3D garment reconstruction.

%% file: figures/fig_dataset_gallery.tex
\begin{figure*}[htbp]
  \centering
  \includegraphics[width=.92\linewidth]{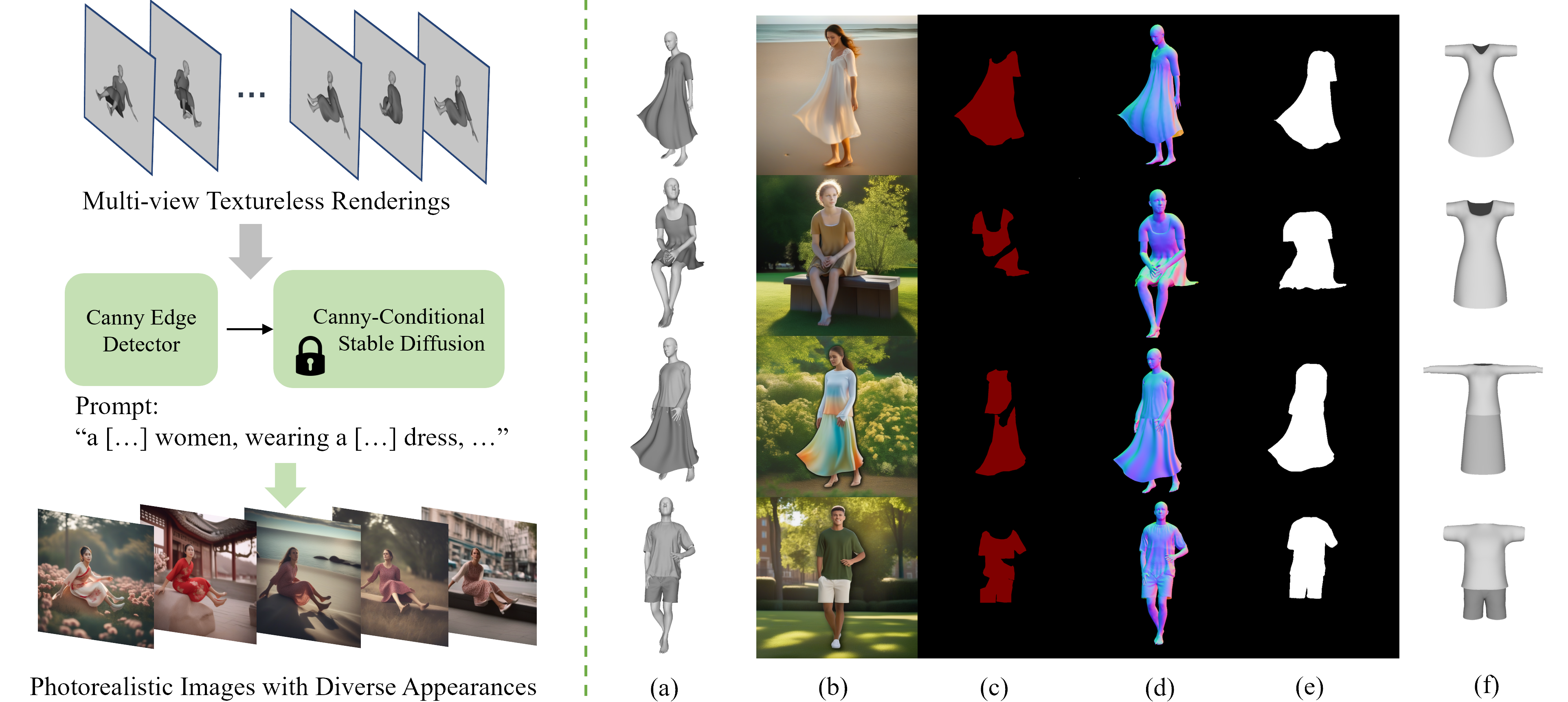}
  \caption{Left: Our novel strategy for generating extensive photorealistic paired images. We acquire rendered images of 3D garments with random camera views. These rendered images are processed through Canny-Conditional Stable Diffusion~\cite{rombach2022high,mou2023t2i,zhang2023adding} to produce photorealistic images. Right: (a) The garment sampled from Fine Garment Dataset; (b) The synthesized image; (c) The pixel-aligned mask; (d) The normal map rendered using (a); (e) The garment mask rendered by (a); (f) The counterpart T-pose coarse garment of (a). In Sec.~\ref{sec:method}, (b, f) is used to train the coarse garment estimator, while (b,c,d) is adopted to train the normal estimator. (d, e, a) is utilized to train the fine garment estimator and the geometry-aware boundary predictor. Synthesized images courtesy of Stable Diffusion.
  }
  \label{fig:data_gallery}
\end{figure*}

%% file: section/4_method.tex
\input{figures/fig_method_pipeline}
\section{Method}
\label{sec:method}

As shown in Fig.~\ref{fig:method_pipeline}, given an RGB image, our approach initially estimates the coarse explicit garment shape $M_P$ (Sec.~\ref{sec:method_ce}). Geometric details are recovered from the implicit function with the assistance of the normal map to obtain a fine garment mesh with a closed boundary $M_I$ (Sec.~\ref{sec:method_fi}). Next, our method combines the 2D image and the 3D fine garment to predict the garment boundary (Sec.~\ref{sec:method_gb}). Finally, we fit the coarse shape with the fine garment mesh by aligning the 3D boundaries to generate the target garment mesh $M_F$ with an open boundary (Sec.~\ref{sec:method_fit}).


\subsection{Coarse Explicit Garment Estimation}
\label{sec:method_ce}
\subsubsection{Unposed Coarse Garment Inference} 

\paragraph{Garment Blendshape Construction} Various current works~\cite{jiang2020bcnet,patel20tailornet,luo2023rabit} demonstrate that linear statistical models are able to represent the basic geometries of diverse shapes. 
Inspired by~\cite{loper2015smpl,jiang2020bcnet}, we utilize PCA to parameterize our unposed (i.e. T-posed) coarse garments by,
\begin{equation}
G({{\alpha}}) = \mathbf{T}_{{g}} 
+ B_{{g}}(\alpha),
\label{math:G}
\end{equation}
where $G(\alpha)$ denotes the statistical garment model, which is worn on top of SMPL's mean shape. $\mathbf{T}_{{g}} \in \mathbb{R}^{N_{{G}} \times 3}$ is a garment template with $N_{{G}}$ vertices. We define an independent T-posed garment template $\mathbf{T}_{{g}}$ for each garment category.
$B_{{g}}(\alpha) \in \mathbb{R}^{N_{{G}} \times 3}$ models the \textit{Garment Shape Blend Shapes} (GSBS) in T-posed space, while $\alpha \in \mathbb{R}^{32}$ is the PCA coefficients that control the GSBS. 

\paragraph{Coarse Garment Estimator} Given an input image, we firstly utilize a lightweight image classifier~\cite{jiang2020bcnet,zhu2020deep,zhu2022registering} to categorize it into one of five common types and select the corresponding statistical garment model. With the selected statistical model, we use a CNN encoder to map the image to the parametric space and obtain T-posed coarse garment $G(\alpha)$ through Eq.~\ref{math:G}.

\subsubsection{Posed Coarse Garment Estimation}
\ \newline 
To model garment's coarse deformation, we extend SMPL's skinning procedure to the pose garment. SMPL incorporates \textit{Body Shape Blend Shapes} (BSBS) and \textit{Body Pose Blend Shapes} (BPBS) to define a T-posed body. Given the shape parameters ($\beta$) and pose parameters ($\theta$), SMPL can generate the T-posed body mesh by,
\begin{equation}
T_{{B}}(\beta, \theta) = \mathbf{T}_{{b}} + B_{{s}}(\beta) + B_{{p}}(\theta),
\end{equation}
where $\mathbf{T}_{{b}} \in \mathbb{R}^{N_{{B}} \times 3}$ is a body template mesh with $N_{{B}}$ vertices. 
$B_{{s}}(\beta) \in \mathbb{R}^{N_{{B}} \times 3}$ denotes the shape-related displacements, while $B_{{p}}(\theta) \in \mathbb{R}^{N_{{B}} \times 3}$ models the pose-dependent correctiveness. Then SMPL uses LBS to pose a rigged template. The mapping can be summarized as the following equation:
\begin{equation}
M_{{B}}(\beta, \theta) = W\left(T_{{B}}(\beta, \theta), J(\beta), \theta, \mathcal{W}\right),
\end{equation}
where $W(\cdot)$ is a skinning function with skinning weights $\mathcal{W} \in \mathbb{R}^{N_{{B}} \times 24}$, joint locations $J(\beta) \in \mathbb{R}^{24 \times 3}$, and pose parameters $\theta$ that rig a T-posed body mesh $T_{{B}}(\beta, \theta)$. 
\paragraph{SMPL Body Estimator}
To pose our garment, we use PyMAF~\cite{pymaf2021} to estimate SMPL parameters from the image, and apply BSBS and BPBS to the T-posed garment by, 
\begin{equation}
T_{{G}}(\alpha, \beta, \theta) = G({{\alpha}}) + \widetilde{B}_{{s}}(G({{\alpha}}), \beta) 
+ \widetilde{B}_{{p}}(G({{\alpha}}), \theta),
\label{math:TG}
\end{equation}
\begin{equation}
\widetilde{B}_{{s}}(G({{\alpha}}), \beta) = w(G({{\alpha}})) B_{{s}}(\beta), 
\end{equation}
\begin{equation}
\widetilde{B}_{{p}}(G({{\alpha}}), \theta) = w(G({{\alpha}})) B_{{p}}(\theta),
\label{math:gbp}
\end{equation}
where $G({{\alpha}})$ is the T-pose garment obtained by Eq.~\ref{math:G}. $\widetilde{B}_{{s}}(\cdot) \in \mathbb{R}^{N_{{G}} \times 3}$ and $\widetilde{B}_{{p}}(\cdot) \in \mathbb{R}^{N_{{G}} \times 3}$ are the corresponding garment displacements influenced by the BSBS and BPBS of the body, respectively. $w(\cdot) \in \mathbb{R}^{N_{{G}} \times N_{{B}}}$ is a weighted matrix that can be computed by searching the K-Nearest Neighbors (KNN) body vertices for each garment vertex~\cite{jiang2020bcnet,peng2021animatable}.

\paragraph{Posed Coarse Garment Modeling} Then we transfer SMPL's LBS to $T_G(\cdot)$ to obtain the posed garment $M_P(\cdot)$ by
\begin{equation}
M_P(\alpha, \beta, \theta) = W\left(T_{{G}}(\alpha, \beta, \theta), J(\beta), \theta, \widetilde{\mathcal{W}}\right),
\label{math:MP}
\end{equation}
\begin{equation}
\widetilde{\mathcal{W}} = w(G({{\alpha}})) \mathcal{W},
\end{equation}
where $\widetilde{\mathcal{W}} \in \mathbb{R}^{N_{{G}} \times 24}$ represents the garment skinning weights extended from SMPL with the same weighted matrix $w(\cdot)$ in Eq.~\ref{math:gbp}.



\subsection{Fine Implicit Garment Reconstruction}
\label{sec:method_fi}
To generate the fine implicit garment field, we first obtain the garment mask and the normal map of the input image (\textit{Please refer to the supplementary materials for details about mask extraction and our normal estimator}).
Then we apply an Hourglass filter~\cite{saito2019pifu} to extract the image feature from the input normal map. The 3D point $p$ is projected to 2D image coordinate by camera projection $\pi(\cdot)$ to exact pixel-aligned local image feature $I_F(p) = F(\pi(p))$. Then we define an implicit function $f$ for arbitrary point $p$ in 3D space as,
\begin{equation}
    f(F(\pi(p)),z(p)) = s:s \in (0, 1),
\end{equation}
where $s$ denotes the occupancy of $p$, and $z(p)$ is the depth in the camera coordinate space. $f(\cdot)$ is designed as MLPs to decode the occupancy status of $p$. $L_1$ loss is chosen to measure the error between the predicted occupancy and the ground truth during training. To compute the occupancy, we use MeshLab's close hole operation~\cite{cignoni2008meshlab} to create a closed topology for each garment.

\subsection{Geometry-aware Boundary Prediction} 
\label{sec:method_gb}
Garment boundaries are thin 3D curves that are challenging to capture with implicit functions. Inspired by ReEF~\cite{zhu2022registering}, we adopt a cylinder structure to represent the garment boundary. To obtain garment boundaries, a straightforward approach is to use 2D image cues to regress the boundary cylinders. However, relying solely on 2D pixel-aligned features suffers from depth ambiguity, leading to inconsistent 3D results. To address the ambiguity, we integrate 2D clues and 3D geometry-aligned features to enhance global boundary shape alignment (see Fig.~\ref{fig:method_pipeline}). We utilize the previous pixel-aligned image feature $F(\pi(p))$ as 2D clues. To produce geometry-aware features from the fine garment $M_I$, we employ a triplane encoder $\psi_{enc}$ to obtain 3D features aligned with three axis-aligned orthogonal planes. Specifically, point clouds sampled from $M_I$ are projected onto the triplane, and a 3D-aware UNet $\psi_{enc}$ is used to obtain high-level triplane feature maps~\cite{liu2024lasa}.
Then we query any 3D position $p \in \mathbb{R}^3$ by projecting it onto each feature plane, retrieving three corresponding feature vectors $G_F(p) = (F_{xy}, F_{xz}, F_{yz})$ via bilinear interpolation. A small MLP-based decoder $\psi_{dec}$ is used to interpret the aggregated concatenated 2D pixel-aligned and 3D triplane features as 3D boundary fields. For an arbitrary 3D point $p$, its occupancy value $o_i$ to the $i$-th boundary is computed as,
\begin{equation}
    f_i(F(\pi(p)),F_{xy}, F_{xz}, F_{yz}) = o_i:o_i \in (0, 1),
\end{equation}
where we model each garment boundary as an implicit cylinder to compute the ground-truth occupancy. $L_1$ loss is used to measure the error between the predicted occupancy and the ground truth.

\subsection{3D Garment Shape Registration}
To obtain the target garment mesh $M_F$, we first establish the boundary correspondence between the boundaries of the coarse garment and the predicted boundary cylinders for registration, as the boundaries possess prominent geometrical features of the garment shape. We fit the coarse mesh boundary strip to the predicted 3D boundary cylinders by minimizing the objective function:
\begin{equation}
    L_{boundary} = \lambda_{c}L_{c} + \lambda_{lap}L_{lap} + \lambda_{edge}L_{edge} + \lambda_{normal}L_{normal},
\end{equation}
where $L_c$ is the Chamfer loss~\cite{ravi2020accelerating} that restricts the positions of boundary mesh vertices; $L_{lap}$, $L_{edge}$ and $L_{normal}$ are Laplacian Smooth, Edge Length and Normal Consistency regularizers~\cite{ravi2020accelerating}, respectively. After fitting the coarse mesh boundary strip, the deformed boundary strip is generated to guide the registration of the target garment mesh. Then, we utilize non-rigid ICP~\cite{amberg2007optimal} to register the coarse garment template to the target garment mesh under the constraints of:
\begin{equation}
    L_{nicp} = \lambda_{d}L_{d} + \lambda_{b}L_{b} + \lambda_{s}L_{s} + \lambda_{reg}L_{reg},
\end{equation}
\begin{equation}
    L_{reg} = \lambda_{lap}L_{lap} + \lambda_{edge}L_{edge} + \lambda_{normal}L_{normal},
\end{equation}
where $L_{d}$ penalizes the distance between the deformed garment template and the ground truth. $L_b$ is the landmark cost between the coarse mesh boundary strips and the deformed boundary strips, while the stiffness term $L_{s}$ penalizes differences between the transformation matrices assigned to neighboring vertices. Different from the original non-rigid ICP, we incorporate the mesh regularization term $L_{reg}$~\cite{ravi2020accelerating} to stabilize the registration process.



\label{sec:method_fit}

%% file: figures/fig_method_pipeline.tex
\begin{figure*}[htbp]
  \centering
  \includegraphics[width=.95\linewidth]{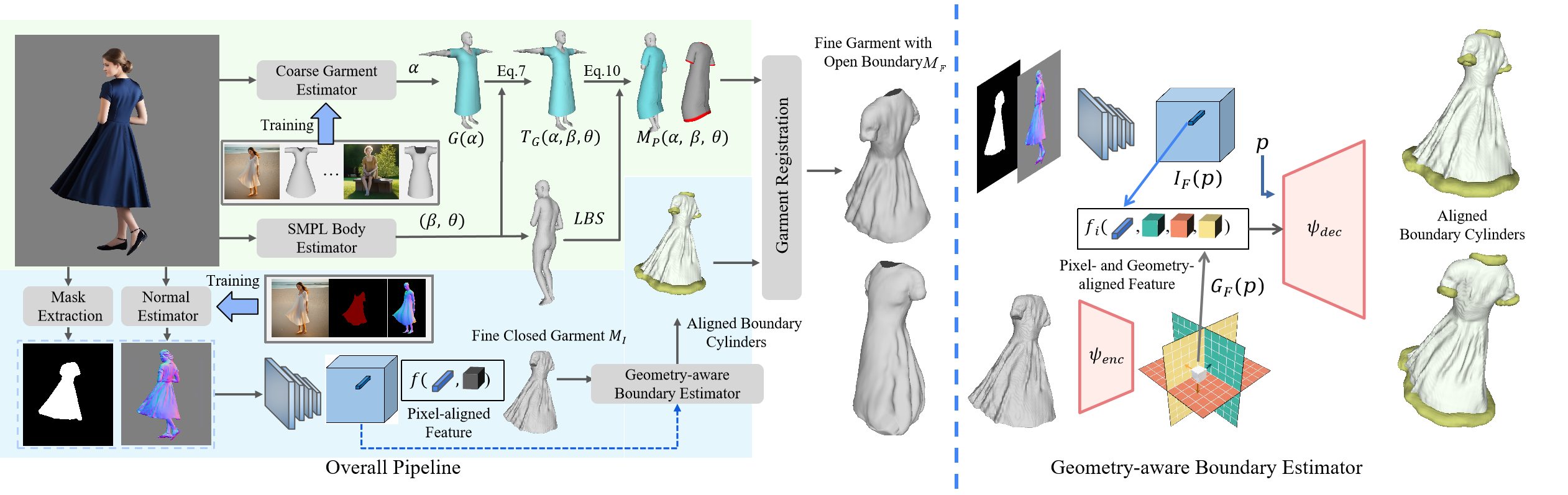}
  \caption{The pipeline of our proposed method. Given an RGB image, our method first estimates the T-pose garment shape $G({{\alpha}})$ (Eq.~\ref{math:G}) and computes its pose-related deformation $M_P(\alpha, \beta, \theta)$ with the help of the predicted SMPL body (Eq.~\ref{math:TG}, Eq.~\ref{math:MP}). Then a pixel-aligned network is used to reconstruct implicit fine garment $M_I$ and the geometry-aware boundary estimator is adopted to predict the garment boundary. Finally, we register $M_P(\cdot)$ to $M_I$ to obtain the final mesh $M_F$, which has fine topology and open-boundaries. Images courtesy of Stable Diffusion.}
  \label{fig:method_pipeline}
\end{figure*}

%% file: section/5_experiment.tex
\input{tables/table_compare}
\input{tables/table_ablation_boundary}
\input{tables/table_ablation_shape}

\section{Experiments}
\label{sec:experiments}

In our experiments, we trained our method and all compared methods using our synthetic dataset (as shown in Fig.~\ref{fig:data_pipeline} and Fig.~\ref{fig:data_gallery}), allocating 80\% for training and reserving the remaining 20\% for testing. To evaluate reconstruction quality, we employ commonly used metrics~\cite{mescheder2019occupancy} for quantitative comparisons, including Chamfer Distance, Normal Consistency, and Intersection over Union (IoU). Quantitative comparisons were conducted on the test-set of our synthetic data, while qualitative comparisons were performed on in-the-wild images.
Fig.~\ref{fig_teaser} and Fig.~\ref{fig:result_gallery} present our representative results. Please refer to our supplementary materials for more results and implementation details.

\paragraph{Comparison Study.} We compare our method with the state-of-the-art single-view garment reconstruction methods, i.e., BCNet, ClothWild, DeepFashion3D and ReEF, both quantitatively and qualitatively. 
Tab.~\ref{tab:compare} shows the quantitative comparisons. Our proposed method achieves the best scores against the baselines. Fig.~\ref{fig:exp_compare} provides qualitative comparisons given the input of in-the-wild images. LBS-based methods (e.g., BCNet, ClothWild) only capture coarse deformations caused by pose and often neglect surface details. Although BCNet utilizes a displacement network to model garment deformations and geometric details, it is challenging for a simple MLP-based network to regress a larger number of vertex offsets. Using only global features makes it inefficient for DeepFashion3D to obtain accurate boundaries, resulting in poor reconstruction quality. Although ReEF performs well on simple poses and simple clothing deformations by leveraging pixel-aligned features, it presents artifacts on garments with complex human poses and garment deformations. Our method demonstrates proficiency in capturing both large garment deformations and geometric details.

\paragraph{Ablation Study on Boundary Prediction.} 
Tab.~\ref{tab:ablation_boundary} shows the quantitative comparisons between ReEF and our method on boundary field prediction. Our proposed method achieves better scores. Fig.~\ref{fig:exp_abl_boundary} provides qualitative comparisons on garment boundary reconstruction. As noted, relying solely on 2D pixel-aligned features makes ReEF fail to predict accurate boundaries with complex poses and deformations, resulting in discontinuous boundaries. Our geometry-aware boundary prediction excels in reconstructing complex garment boundaries that are well-aligned with the garment shape.

\paragraph{Ablation Study on Data.}
We verify the significance of our data by training our method on both: 1) ReEF's dataset; and 2) our \dataName. As shown in Fig.~\ref{fig:exp_abl_data} and Tab.~\ref{tab:ablation_shape}, the model trained with our data achieves the best results, indicating that our data enhances the network's generalization in reconstructing in-the-wild images.

\paragraph{Ablation Study on Coarse Garment Estimation.}
To demonstrate the significance of using our dataset in building the garment parametric model, we conduct an ablation study on different methods to obtain coarse garments. Apart from using our parametric model and estimator, there is an alternative strategy: cropping a part of the mesh from a posed SMPL body, as used in DeepFashion3D and ReEF. Fig.~\ref{fig:exp_abl_coarse} and Tab.~\ref{tab:ablation_shape} present the comparisons between our method and the ablated strategies. Our method is superior in estimating a more reasonable coarse garment. The registered results show that a good coarse initialization significantly stabilizes the registration process.

\paragraph{Ablation Study on Implicit Representation.}
Apart from registering coarse garments to fine garments, another strategy for obtaining open-boundary meshes is to use UDF (Unsigned Distance Field). However, UDF encounters two problems (Fig.~\ref{fig:exp_abl_implicit_udf}, Tab.~\ref{tab:ablation_shape}): 1) Although some methods~\cite{guillard2022udf} can extract open-boundary meshes from UDF, the quality is poor and may result in unexpected open regions and incomplete meshes. Garment registration is still required to achieve fine topology. 2) The regression problem with UDF is more challenging to converge than classification, resulting in inferior surface details compared to the occupancy field.

%% file: tables/table_compare.tex

\begin{table}[htbp]
    \centering
    \caption{\textbf{Quantitive comparison between our method with others}.}
    \resizebox{0.49\textwidth}{!}{
        \begin{tabular}{c|ccccc}
          Method & BCNet & ClothWild & Deep Fashion3D & ReEF & Ours \\
        \midrule
        Chamfer Distance $\downarrow$ & 18.742 & 16.136 & 17.159 & 11.357 & \textbf{7.825} \\
        Normal Consistency $\uparrow$ & 0.781 & 0.812 & 0.793 & 0.838 & \textbf{0.913} \\
        \end{tabular}
    }
    \label{tab:compare}
\end{table}

%% file: tables/table_ablation_boundary.tex

\begin{table}[htbp]
    \centering
    \caption{\textbf{Quantitative comparison between our method and alternative strategies for predicting garment boundary}.}
    \resizebox{0.39\textwidth}{!}{
        \begin{tabular}{c|ccc}
        Method & Chamfer Distance $\downarrow$ & Normal Consistency $\uparrow$ & IoU $\uparrow$ \\
        \midrule
        ReEF & 16.428 & 0.809 & 55.425 \\
        Ours  & \textbf{10.571} & \textbf{0.862} & \textbf{69.775} \\
        \end{tabular}
    }
    \label{tab:ablation_boundary}
\end{table}

%% file: tables/table_ablation_shape.tex
\begin{table*}[htbp]
    \centering
    \caption{\textbf{Quantitative comparison between our method and alternative strategies}.}
    \resizebox{0.82\textwidth}{!}{
        \begin{tabular}{c|c|c|ccc|c}
        \multirow{3}{*}{Method} & \multicolumn{5}{c|}{\textit{Ablation Study on}} & \multirow{3}{*}{Ours} \\
        & \multicolumn{1}{c|}{\textit{Data}} & \multicolumn{1}{c|}{\textit{Coarse Garment Estimation}} & \multicolumn{3}{c|}{\textit{Implicit Representation}} & \\
          & ReEF's dataset & Crop from SMPL & UDF w/o Registering & UDF w/ Registering & Occupancy w/o Registering & \\
        \midrule
        Chamfer Distance $\downarrow$ & 16.363 & 14.635 & 9.616 & 9.375 & 8.658 & \textbf{7.825}\\
        Normal Consistency $\uparrow$ & 0.805 & 0.823 & 0.841 & 0.848 & 0.851 & \textbf{0.913} \\
        \end{tabular}
    }
    \label{tab:ablation_shape}
\end{table*}

%% file: section/6_conclusion.tex
 \section{Conclusion}
\label{sec:conclusion}
Capturing diversified garment shapes and intricate garment deformations robustly from single RGB images remains difficult due to garment complexity and data scarcity. Our work presents a large-scale 3D garment dataset \dataName, which is extensively annotated at different levels of detail, ranging from coarse stylized garments to deformed models with intricate deformations and fine-grained geometric details. Based on the well-established dataset, we propose a framework for high-quality 3D garment reconstruction from single-view images. The core of our approach is a hierarchical design to recover different levels of garment details, i.e., from pose-independent stylized coarse garments to pose-blended and open-boundary garments with pixel-aligned details. Experiments indicate that our framework is capable of reconstructing garments with various shapes and fine-grained deformations, showcasing its superior generalization ability against state-of-the-art methods.


\paragraph{Limitation.} Although our work provides faithful reconstructed results on a wide range of in-the-wild images, it may fail when reconstructing garments with complex topology: 1) As shown in Fig.~\ref{fig:limitation}(a), although our method is able to reconstruct faithful clothing details, it fails to represent the multi-layer structures present in dresses or skirts. This problem is largely due to the reliance on the single-layer occupancy field and the single-layer garment parametric model, which are unable to capture multi-layered structures. One possible solution is to design a new representation that effectively supports the reconstruction of garments with multi-layer structures; 2) As shown in Fig.~\ref{fig:limitation}(b), our method struggles to accurately reconstruct dresses or skirts with slits. This issue primarily stems from the limited representation of such features in our current dataset. The lack of sufficient examples of slits in the training data restricts the model's ability to generalize and accurately reconstruct these specific structures. A potential strategy is to expand the dataset by incorporating a broader range of clothing styless, thereby enhancing the model’s capability to handle these intricate features.


%% file: figures/fig_result_gallery.tex
\begin{figure*}[htbp]
  \centering
  \includegraphics[width=0.90\linewidth]{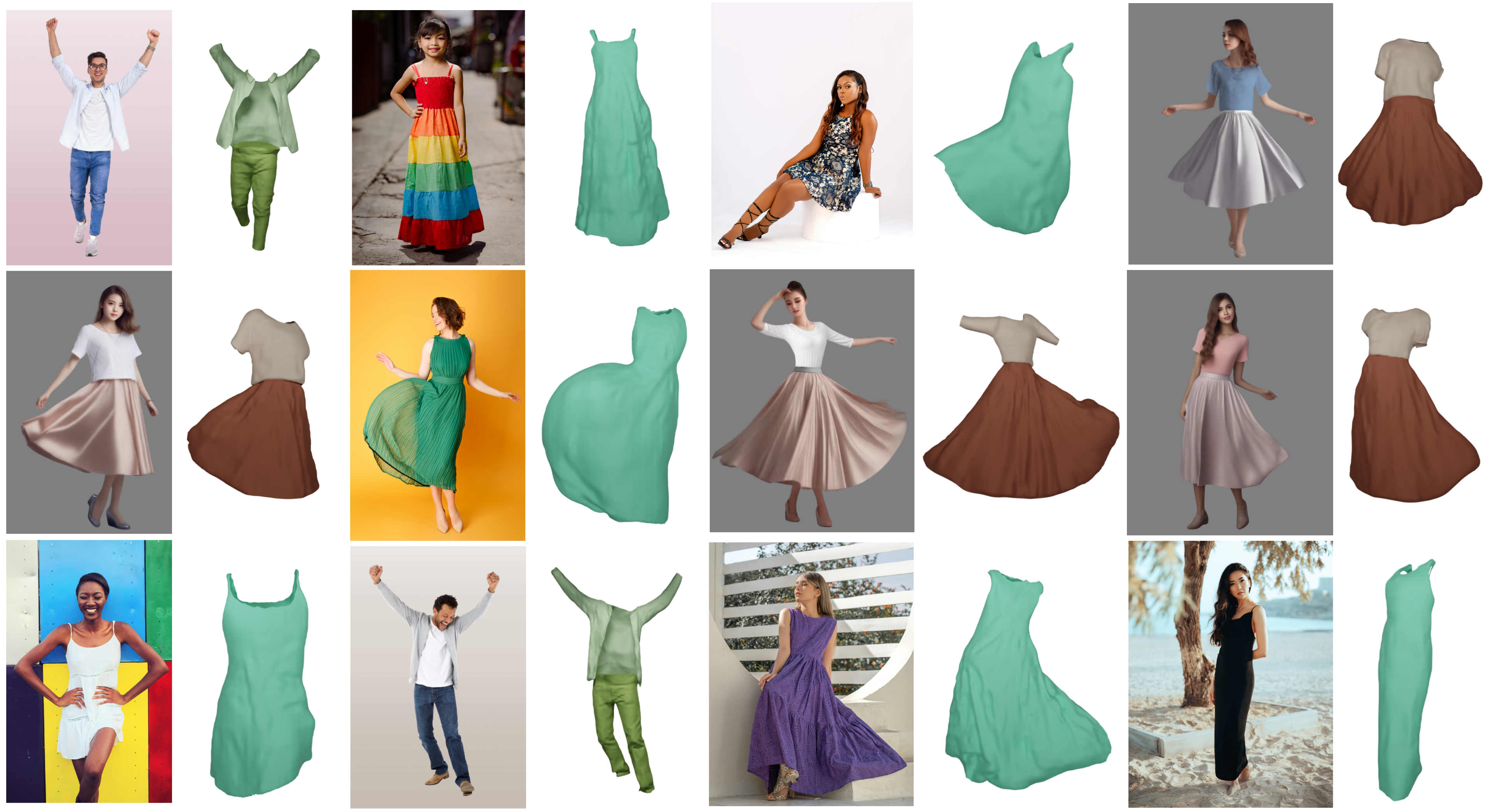}
  \caption{Result gallery of our method. Each image is followed by the reconstructed garment mesh. As illustrated, our method can effectively reconstruct garments with intricate deformations and fine-grained surface details. To support the modeling of folded structures, such as collars, we assembled a repository of diverse real-world collars that were crafted based on our topologically-consistent garments. A lightweight classification network was trained to select the collar that best matches the given image in terms of appearance~\cite{zhu2022registering}. Original images courtesy of licensed photos and Stable Diffusion. The images with a gray background are synthesized, while the rest are licensed photos.}
  \label{fig:result_gallery}
\end{figure*}

%% file: figures/fig_compare.tex
\begin{figure*}[htbp]
  \centering
  \includegraphics[width=0.80\linewidth]{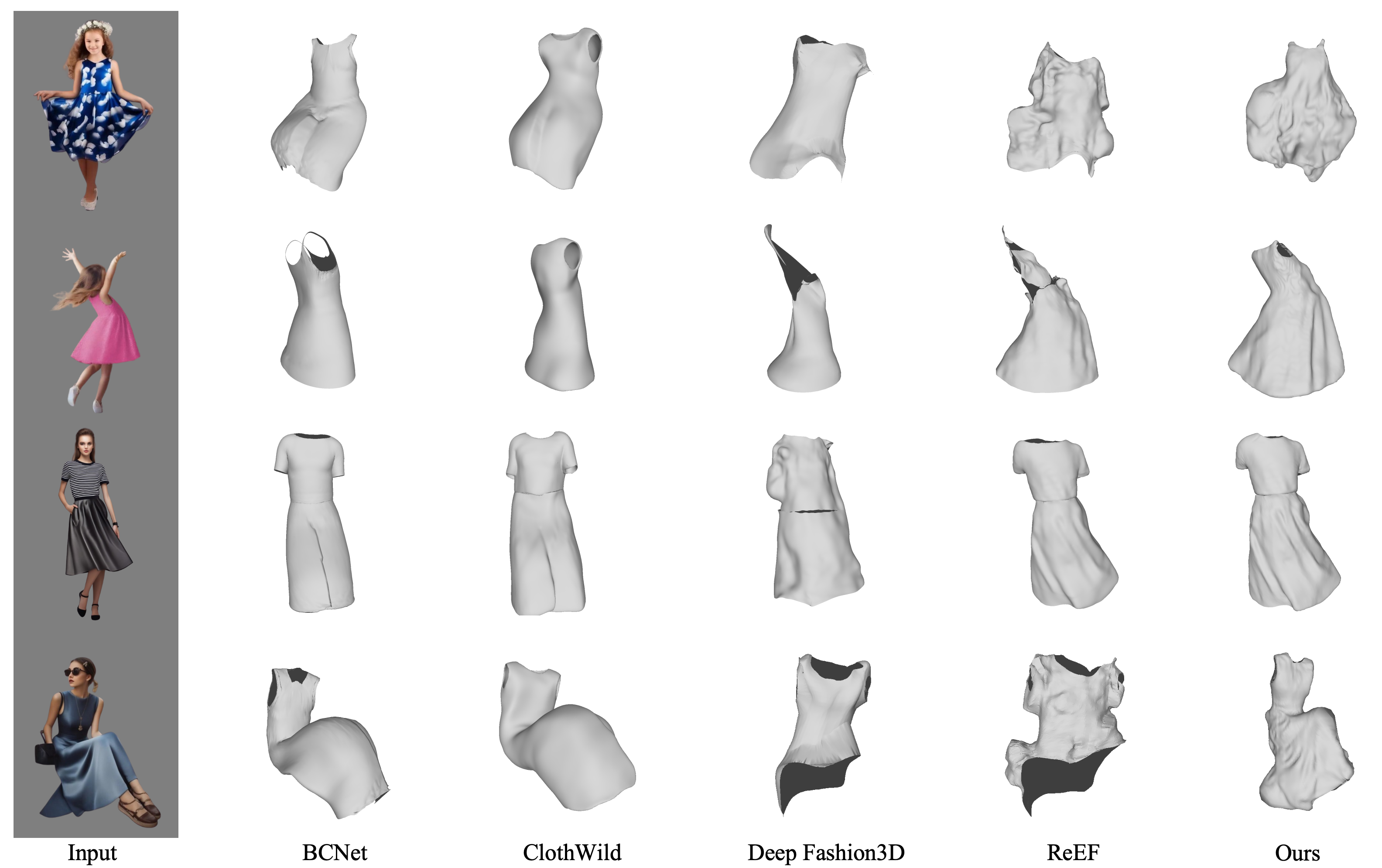}
  \caption{Qualitative comparison between ours and the state of the arts. For each row, the input image is followed by the results generated by BCNet~\cite{jiang2020bcnet}, ClothWild~\cite{moon20223d}, Deep Fashion3D~\cite{zhu2020deep}, ReEF~\cite{zhu2022registering} and our method. Input images courtesy of Stable Diffusion.}
  \label{fig:exp_compare}
\end{figure*}

%% file: figures/fig_ablation_boundary.tex
\begin{figure}[H]
  \centering
  \includegraphics[width=.99\linewidth]{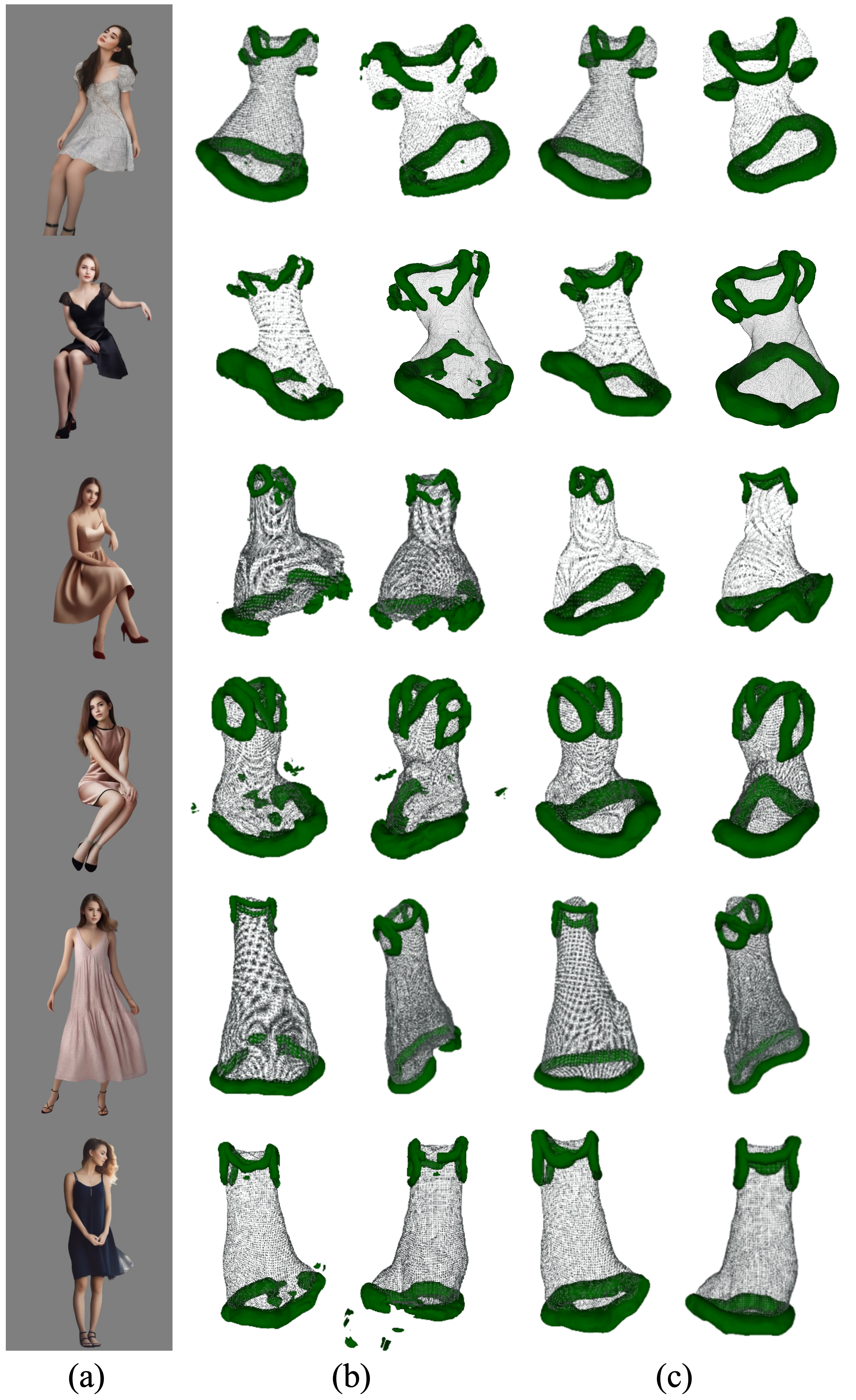}
  \caption{Qualitative comparison between our method and the alternative strategy for predicting garment boundary from in-the-wild images. The input image (a) is followed by the boundaries generated by (b) ReEF's strategy and (c) our geometry-aware estimator. ReEF fails to accurately predict boundaries with complex poses and deformations, leading to discontinuous boundaries. Our geometry-aware boundary prediction outperforms ReEF in reconstructing complex garment boundaries that are well-aligned with the garment shape. Input images courtesy of Stable Diffusion.}
  \label{fig:exp_abl_boundary}
\end{figure}

%% file: figures/fig_alation_data.tex
\begin{figure}[H]
  \centering
  \includegraphics[width=.99\linewidth]{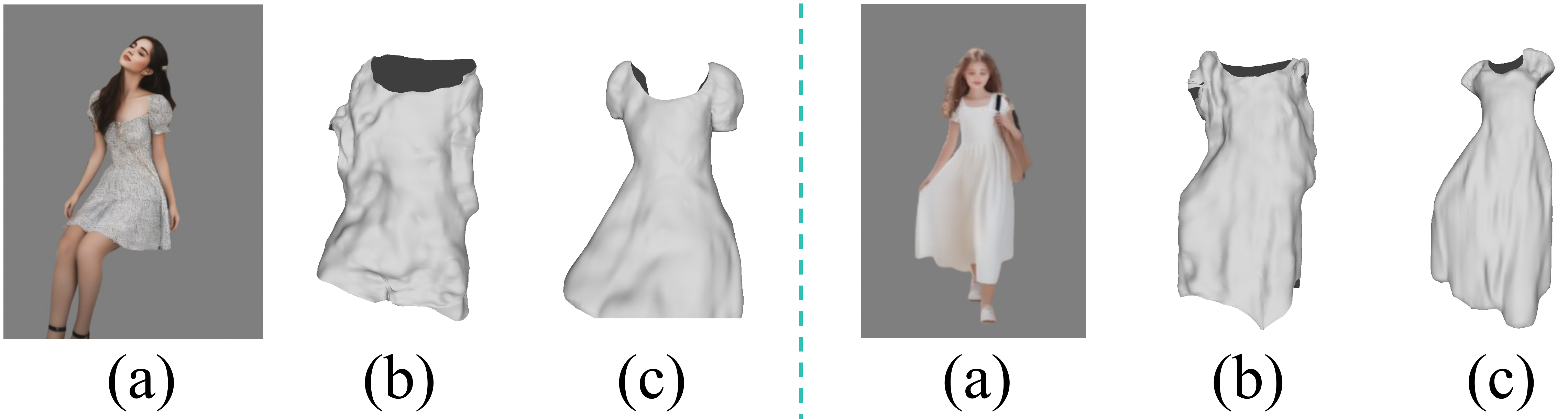}
  \caption{Qualitative comparison on different data. The input image (a) is followed by the results generated by networks trained with (b) ReEF's data and (c) our \dataName. Input images courtesy of Stable Diffusion.}
  \label{fig:exp_abl_data}
\end{figure}

%% file: figures/fig_ablation_coarse.tex
\begin{figure}[H]
  \centering
  \includegraphics[width=.93\linewidth]{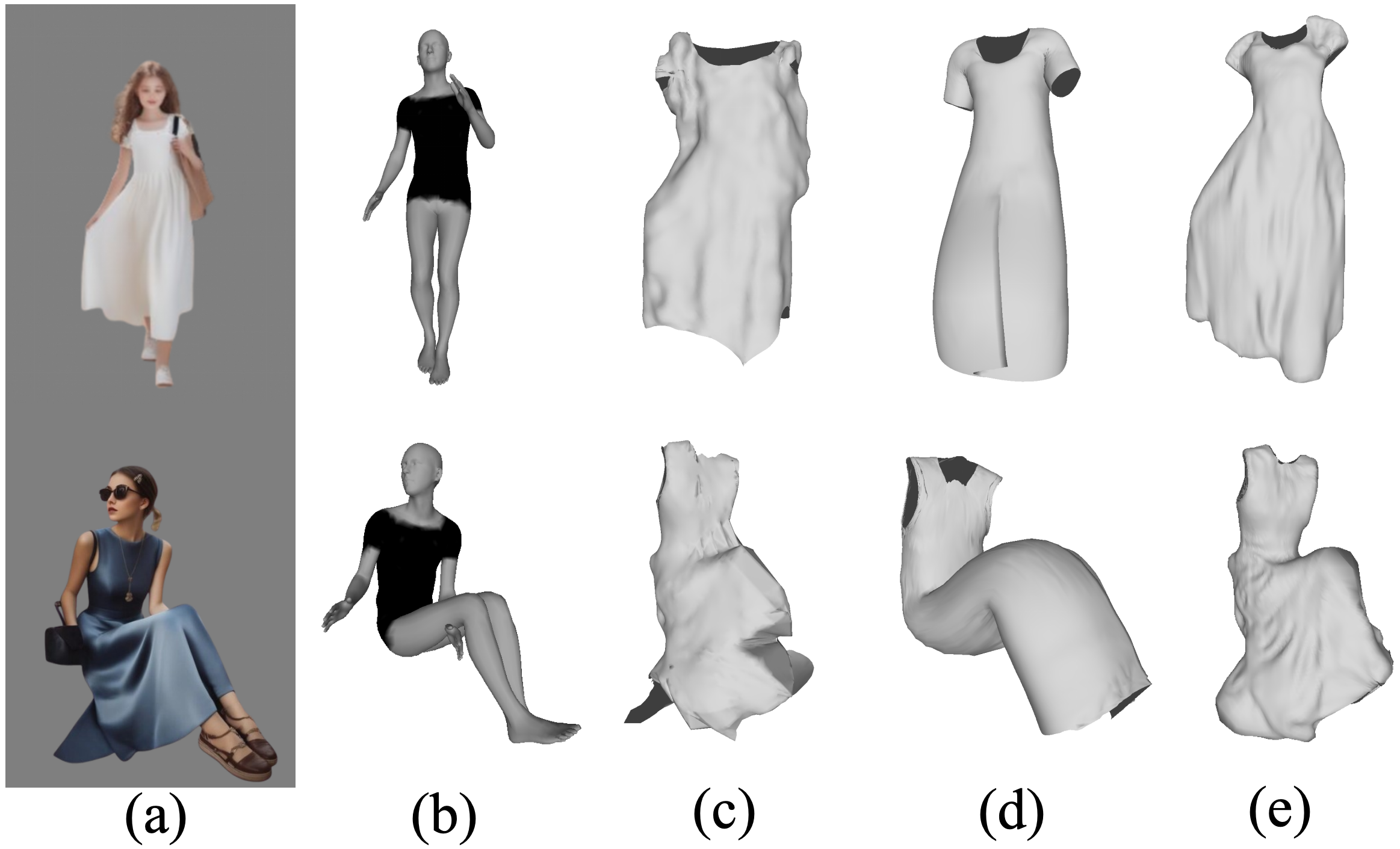}
  \caption{Qualitative comparison between our method and the alternative strategy for obtaining coarse garment template. (a) the input image; (b) the template (black part) cropped from SMPL; (c) the registration result using (b); (d) the coarse garment estimated by our coarse garment estimator; and (e) the registration result using (d). Input images courtesy of Stable Diffusion.}
  \label{fig:exp_abl_coarse}
\end{figure}

%% file: figures/fig_ablation_implicit_udf.tex
\begin{figure}[H]
  \centering
  \includegraphics[width=.90\linewidth]{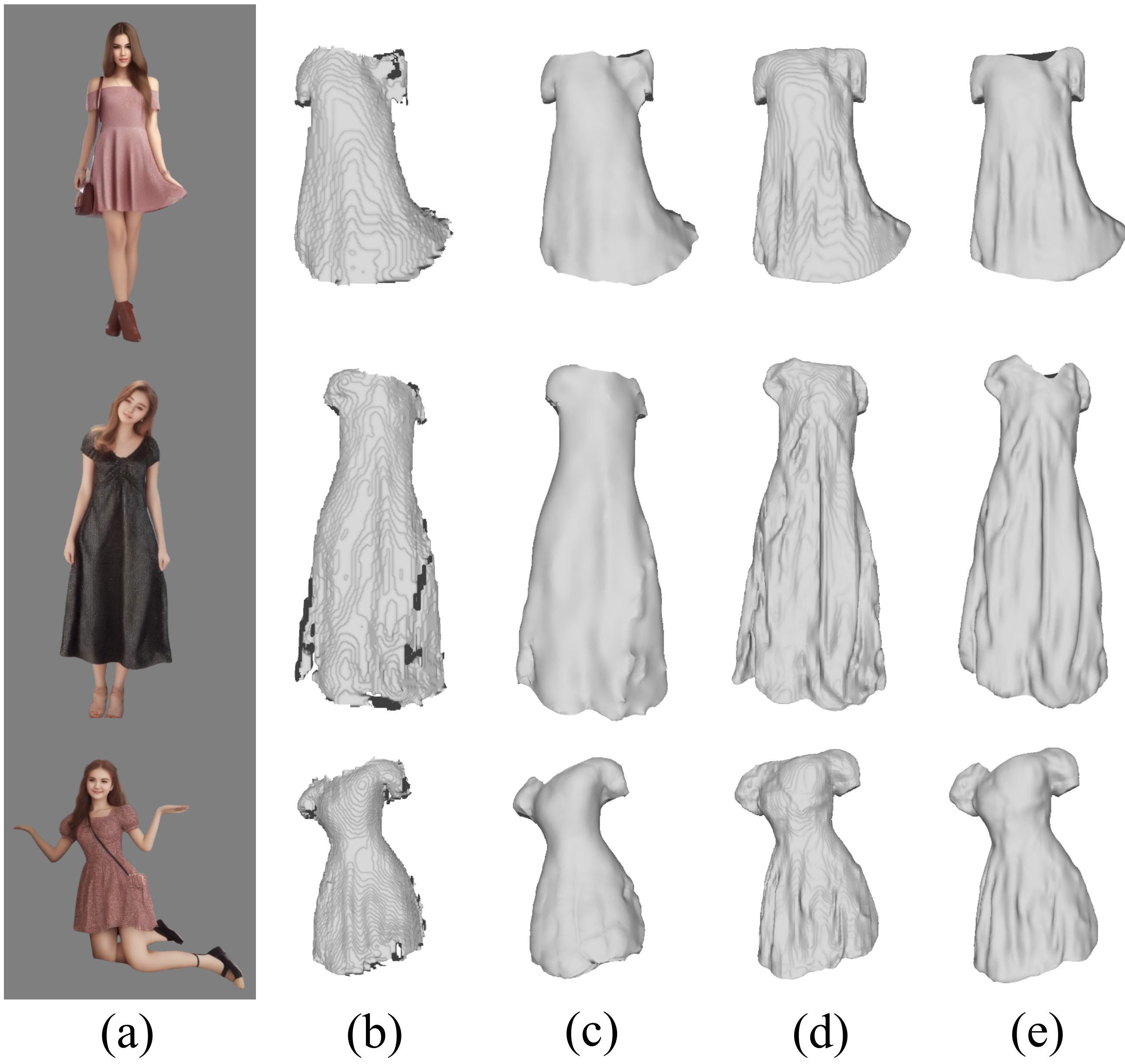}
  \caption{Qualitative comparison on different representation. The input image (a) is followed by the result generated by (b) UDF,  (c) registering to (b), (d) occupancy field and (e) registering to (d). Input images courtesy of Stable Diffusion.}
  \label{fig:exp_abl_implicit_udf}
\end{figure}

%% file: figures/fig_limitation.tex
\begin{figure}[H]
  \centering
  \includegraphics[width=.90\linewidth]{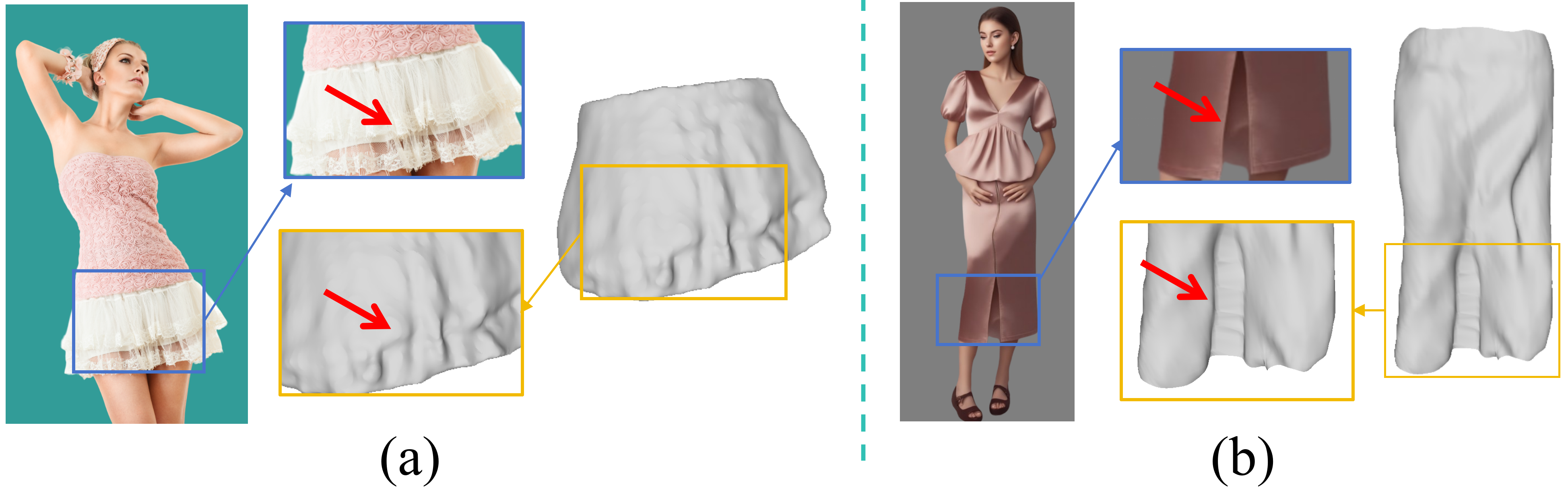}
  \caption{Failure cases. Our framework may struggle to reconstruct garments with complex topology, such as those multi-layered structures (a) or featuring slits (b). Images courtesy of licensed photos and Stable Diffusion.}
  \label{fig:limitation}
\end{figure}

%% file: supp/8_implentation.tex
\section{More Results and Implementation}
\paragraph{More Results.} We report more results on challenging loose cloth reconstruction in Fig.~\ref{fig:result_2}, Fig.~\ref{fig:result_3}, Fig.~\ref{fig:result_4} and Fig.~\ref{fig:result_1}.

\paragraph{Implementation details} In our implementation, all networks were implemented using PyTorch and trained on an Ubuntu server equipped with four A100 GPUs. Quantitative evaluations and qualitative assessments were also performed on this server. For the coarse garment estimator, the parameter size of our statistical garment model $G(\alpha)$ is 32 (i.e., the length of $\alpha$ is 32). We use ResNet-50 blocks~\cite{marcel2010torchvision,luo2023rabit} to map the input image ($512 \times 512 \times 3$) to $\alpha$ and obtain the T-posed coarse garment through Eq.~4 in the Main Paper. All the data in the Garment Style Database were used to learn $G(\alpha)$. To obtain a powerful estimator, we collected 10,000 image-3D T-pose garment paired data (as shown in Fig.~3(b) and Fig.~3(f) of the Main Paper) for training. The coarse garment estimator is trained for $1000$ epochs using the Adam optimizer with a batch size of $128$ and a learning rate of $3 \times 10^{-4}$. For the SMPL body estimator, we employ the well-established body estimator PyMAF~\cite{pymaf2021,pymafx2023} to predict body shape and pose.

Inspired by the normal-aided approaches~\cite{saito2020pifuhd, xiu2022icon}, we employ the normal map and garment segmentation mask as inputs to accurately carve the garment surface details. In the training stage, the ground-truth normal maps and garment masks of our synthetic data are directly rendered from the 3D garment models of \dataName. Given the ground-truth normal map ($512 \times 512 \times 3$) and the corresponding collected images ($512 \times 512 \times 3$), we train a normal estimator using a U-Net~\cite{ronneberger2015u} network with the following loss:
\begin{equation}
    L_N=L_{pixel}[M] + \lambda_{VGG}L_{VGG}[M],
\end{equation}
where $L_{pixel}$ is a $L_1$ Loss between the ground-truth and predicted normal maps and $L_{VGG}$ is a perceptual loss~\cite{johnson2016perceptual} weighted by $\lambda_{VGG}$. M is the pixel-aligned mask as shown in Fig.~3(c) of the Main Paper. We collect 5,000 paired data (as illustrated in Fig.~3(b,c,d) of the Main Paper to train the normal estimator. To improve the performance of the normal estimator, we also incorporate data from THUman2~\cite{tao2021function4d}.
The normal estimator is trained for $80$ epochs using the Adam optimizer with a batch size of $32$ and a learning rate of $3 \times 10^{-4}$. In the testing stage, the garment masks of in-the-wild images are generated by leveraging the segmentation of SAM~\cite{Kirillov2023SegmentA}, and the normal maps are predicted by our trained estimator.

For the fine garment estimator and the geometry-aware boundary predictor, the input size of the normal map and the garment mask is $512 \times 512 \times 3$. Inspired by~\cite{saito2019pifu}, we utilize an Hourglass filter to extract image features and employ an MLP network to decode the features of each sampled point into an occupancy value. 
The fine garment estimator undergoes 100 epochs of training with the RMSprop optimizer, utilizing a batch size of $16$ and a start-up learning rate of $1 \times 10^{-3}$. The learning rate is reduced by a factor of 10 after epochs $30$, $60$, and $90$.
The geometry-aware triplane features are set to a resolution of $256 \times 256$. The boundary estimator is trained over 100 epochs using the RMSprop optimizer, with a batch size of $12$ and a starting learning rate of $1 \times 10^{-3}$. The learning rate is also reduced by a factor of 10 following epochs $30$, $60$, and $90$.
We construct 5,000 pairs of data for each category using the data synthesis strategy shown in Fig.~2 and Fig.~3 of the Main Paper to train our fine garment estimator and the geometry-aware boundary predictor. Although our experiments were conducted with 5,000 pairs of data, it is important to note that our strategy is capable of synthesizing larger-scale datasets, given sufficient computing resources. For garment shape registration, it is generally better to use different weight schedulers to optimize various types of clothing. Please refer to our code for details.

%% file: supp/7_data.tex
\section{Details of \dataName}
\dataName spans a wide range of 3D garment models, containing 5 common garment categories, i.e., dress, skirt, coat, top, and pant. Tab.~\ref{tab:dataset} shows the statistical data for each garment category. The total size represents the number of garments created by artists, not the size of garments our strategy can synthesize. We have recruited eight professional artists to create corresponding 3D garments using Blender based on the collected reference images. The topological consistency of garments enables us to generate new samples by interpolating between two garments within each database. Theoretically, we can generate more garments than the product of the sizes of the individual databases. All eight artists are required to craft 3D models by deforming the predefined template mesh. Each artist possesses over five years of modeling experience, and on average, each garment takes around average 25 minutes to complete. Fig.~\ref{fig:data_template} illustrates the predefined template meshes for each category. Fig.~\ref{fig:LOD_0}, Fig.~\ref{fig:LOD_1}, Fig.~\ref{fig:LOD_2} and Fig.~\ref{fig:LOD_3} respectively illustrate our four datasets: 1) Garment Style Database; 2) Local Detail Database; 3) Garment Deformation Database and 4) Fine Garment Dataset.

\input{supp_tables/table_dataset}
\input{supp_figures/fig_data_template}

\input{supp_figures/fig_deformation_craft}

\paragraph{Garment Deformation Crafting.} As shown in Fig.~\ref{fig:deformation_craft}, we first collect real images of clothed humans with diverse poses, garment styles, and deformations from the Internet, covering the 5 garment categories. Secondly, we employ PyMAF~\cite{pymaf2021} to estimate the human shape $\beta$ and pose $\theta$ from the images and discard the inaccurate estimation results manually. Thirdly, we recruited 8 artists to construct 3D clothing models manually to match the reference images as much as possible, following the specific procedure below: 1) The artists are required to create the coarse \textbf{T-pose Garments} according to the Collected Images, by deforming the predefined category-specific templates to match T-pose SMPL meshes generated by PyMAF; 2) Then, the SMPL's Linear Blend Skinning (LBS) is extended to the T-pose garments programmatically to capture garment deformations resulting from human poses, obtaining the \textbf{Posed Garments}; 3) Finally, the artists are asked to deform the Posed Garments to create the final \textbf{Crafted Garments}, ensuring that deformations match the collected images as closely as possible. In-the-wild images naturally capture the complex real-world physical conditions that occur in a single snapshot. By basing manual modeling on reference images, our data encompass diverse clothing-states observed in real-world scenarios. Note that \textbf{Posed Garments} represent the shape of garments after being affected by human pose, while \textbf{Deformed Garments} (i.e, Crafted Garments) capture the state of garments affected by complex factors (not only affected by pose, but also by other complex environmental factors, such as garment-environment interactions and external forces like wind).

\paragraph{Notation table.} Tab.~\ref{tab:notation} provides a summary of the notations used in the Main Paper.

%% file: supp_tables/table_dataset.tex
\begin{table}[htbp]
    \centering
    \caption{Data statistics for each basic database. The total size refers to the number of garments crafted by artists.}
    \resizebox{0.45\textwidth}{!}{
        \begin{tabular}{l|ccccc}
        Category & Dress & Coat & Skirt & Top & Pant \\
        \midrule
        Garment Style Database & 863 & 760 & 538 & 350 & 358 \\
        Local Detail Database & 86 & 62 & 55 & 38 & 36 \\
        Garment Deformation Database & 622 & 605 & 456 & 582 & 589 \\
        \midrule
        Total & 1,571 & 1,427 & 1,049 & 970 & 983 \\
        \end{tabular}
    }
    \label{tab:dataset}
\end{table}

%% file: supp_figures/fig_data_template.tex
\begin{figure}[htbp]
  \centering
  \includegraphics[width=.88\linewidth]{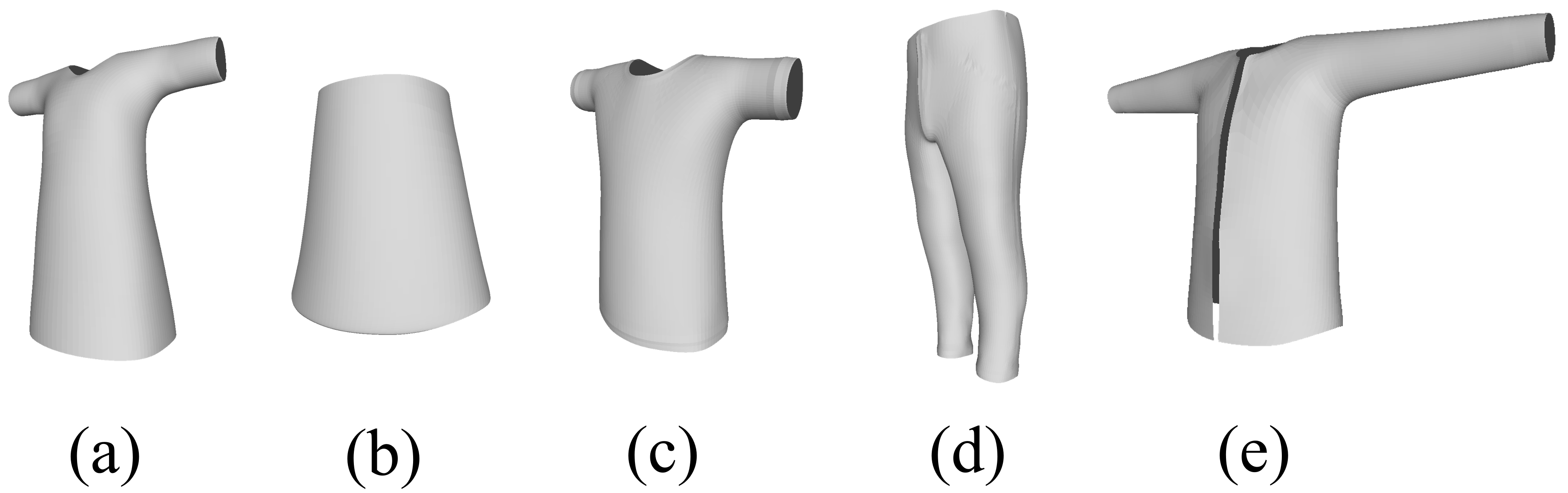}
  \caption{Predefined templates for each garment category, including (a) dress, (b) skirt, (c) top, (d) pant, and (e) coat.}
  \label{fig:data_template}
\end{figure}

%% file: supp_figures/fig_deformation_craft.tex
\begin{figure*}[htbp]
  \centering
  \includegraphics[width=.68\linewidth]{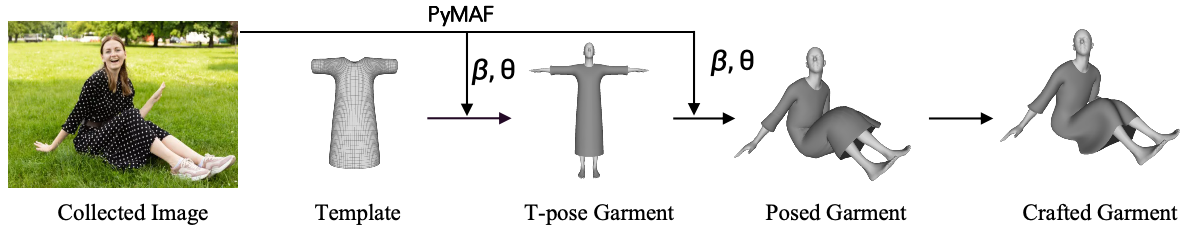}
  \caption{Given a ``Collected Image'', we utilize PyMAF~\cite{pymaf2021,pymafx2023} to estimate SMPL body. Eight artists are then tasked with creating ``T-pose Garment'' shapes by deforming a predefined ``Template'' to match the T-pose body predicted by PyMAF. Then the SMPL's Linear Blend Skinning (LBS) is extended to the T-pose garment to obtain the ``Posed Garment''. Finally, the artists are further instructed to refine the posed garment to get the ``Crafted Garment'' while ensuring that garment deformations closely match the collected images. ``Posed Garment'' represent the shape of clothing influenced by human pose, while ``Crafted Garment'' capture the state of garments affected by various complex factors—not only pose but also other environmental influences, such as garment-environment interactions and external forces like wind.}
  \label{fig:deformation_craft}
\end{figure*}

%% file: supp/9_more_data_more_results.tex
\begin{figure*}[htbp]
  \centering
  \includegraphics[width=.96\linewidth]{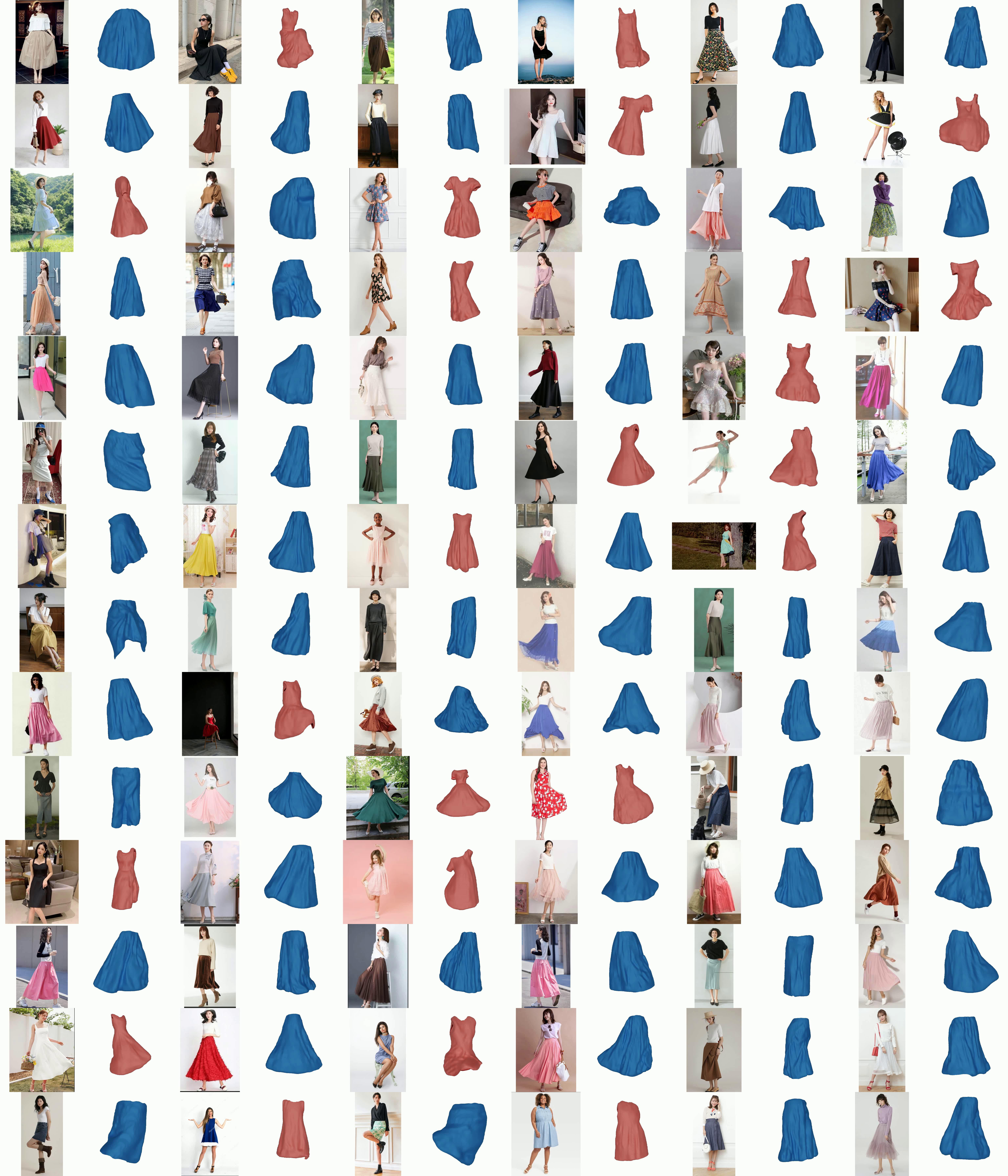}
  \caption{More Results on Loose-fitting Garments.}
  \label{fig:result_2}
\end{figure*}

\begin{figure*}[htbp]
  \centering
  \includegraphics[width=.96\linewidth]{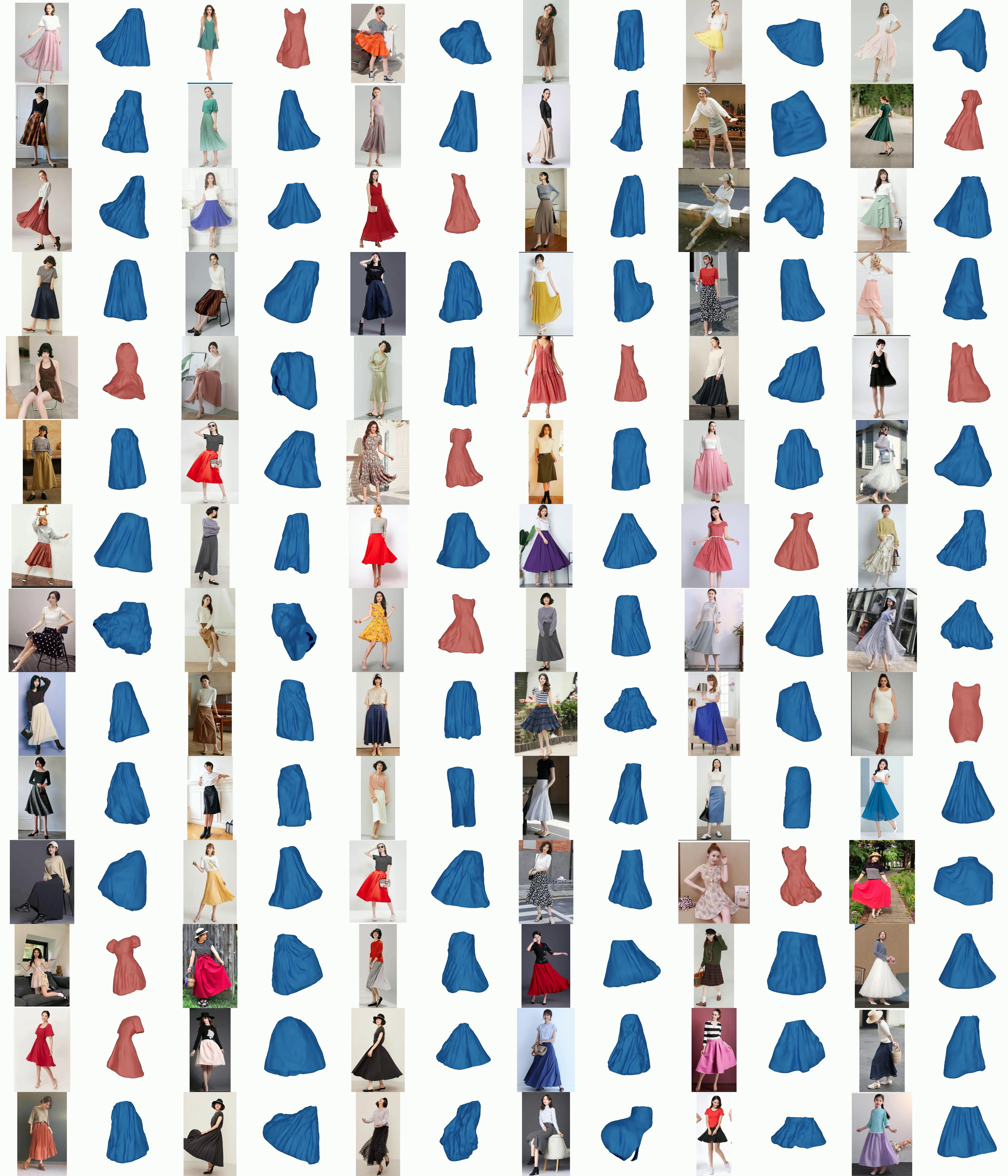}
  \caption{More Results on Loose-fitting Garments.}
  \label{fig:result_3}
\end{figure*}

\begin{figure*}[htbp]
  \centering
  \includegraphics[width=.96\linewidth]{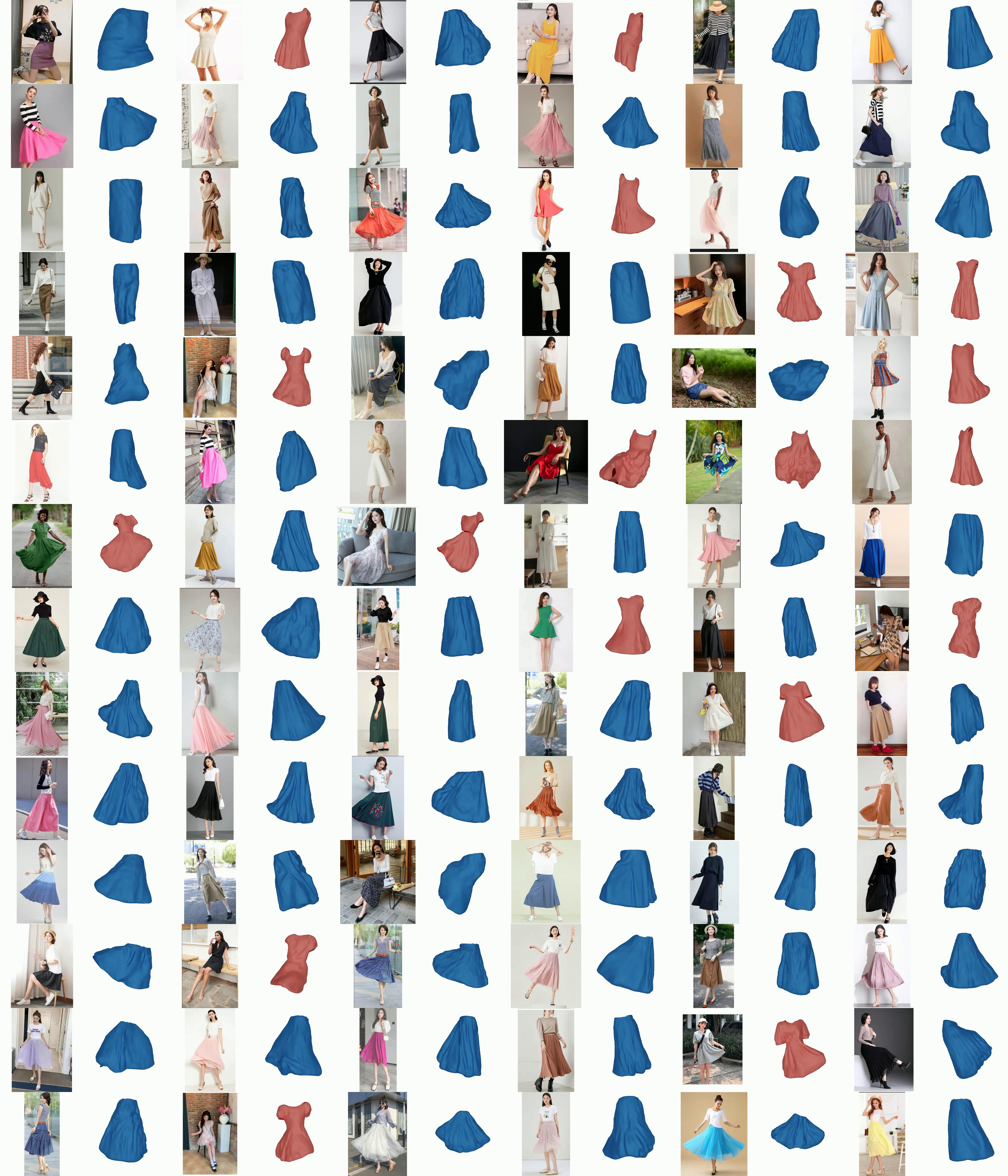}
  \caption{More Results on Loose-fitting Garments.}
  \label{fig:result_4}
\end{figure*}

\begin{figure*}[htbp]
  \centering
  \includegraphics[width=.96\linewidth]{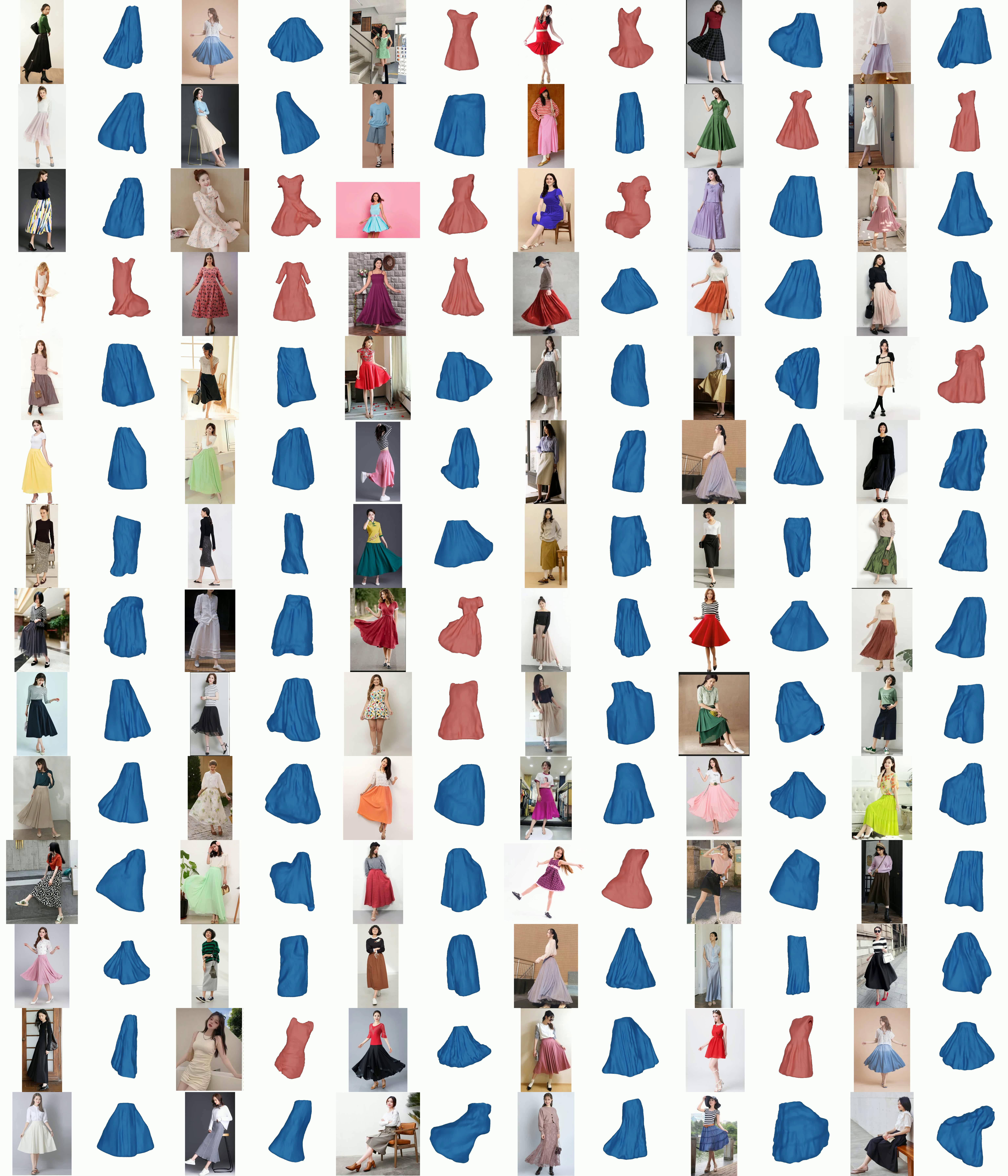}
  \caption{More Results on Loose-fitting Garments.}
  \label{fig:result_1}
\end{figure*}

\begin{figure*}[htbp]
  \centering
  \includegraphics[width=.96\linewidth]{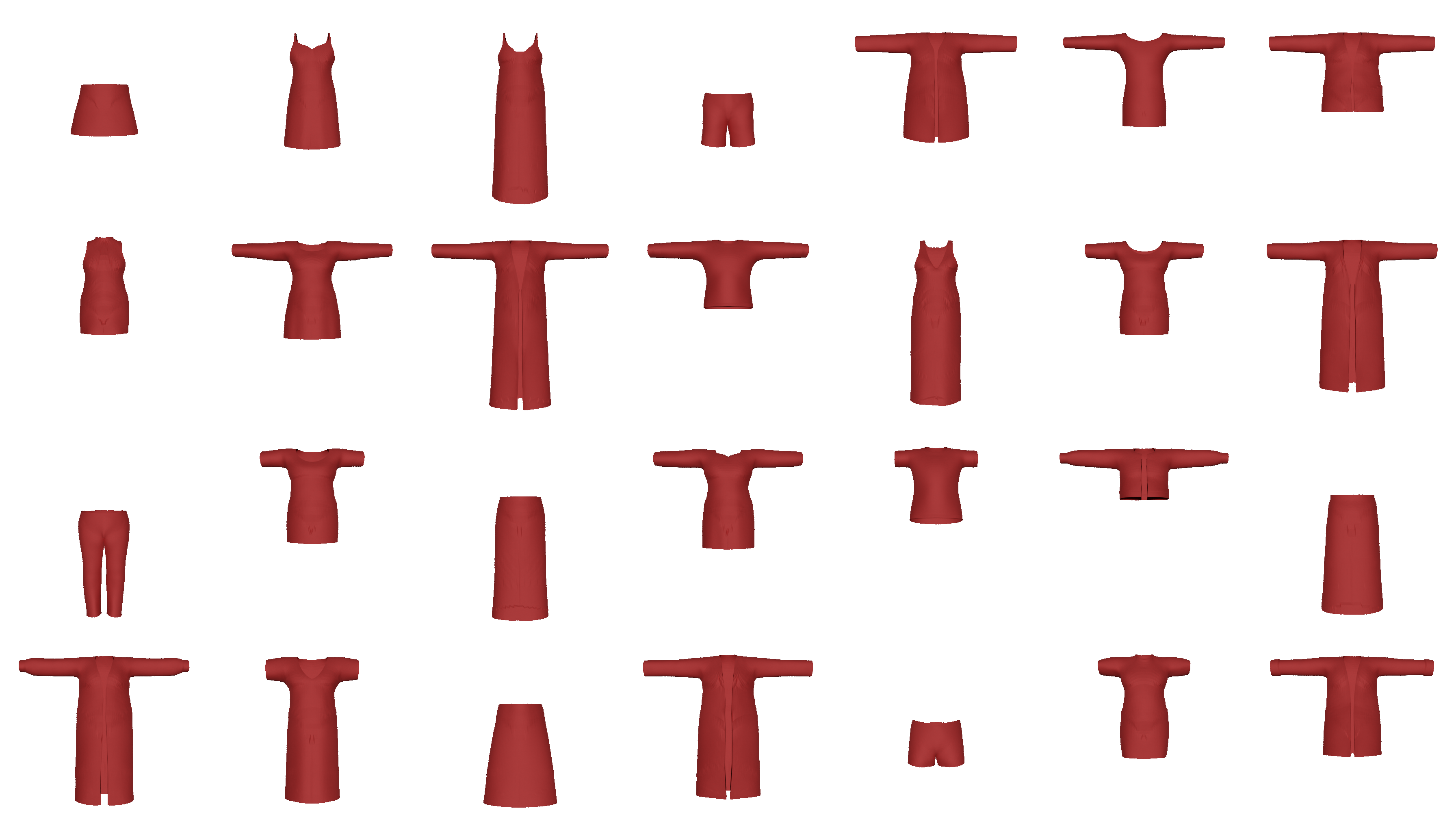}
  \caption{An illustration of our \textbf{Garment Style Database}.}
  \label{fig:LOD_0}
\end{figure*}
\begin{figure*}[htbp]
  \centering
  \includegraphics[width=.96\linewidth]{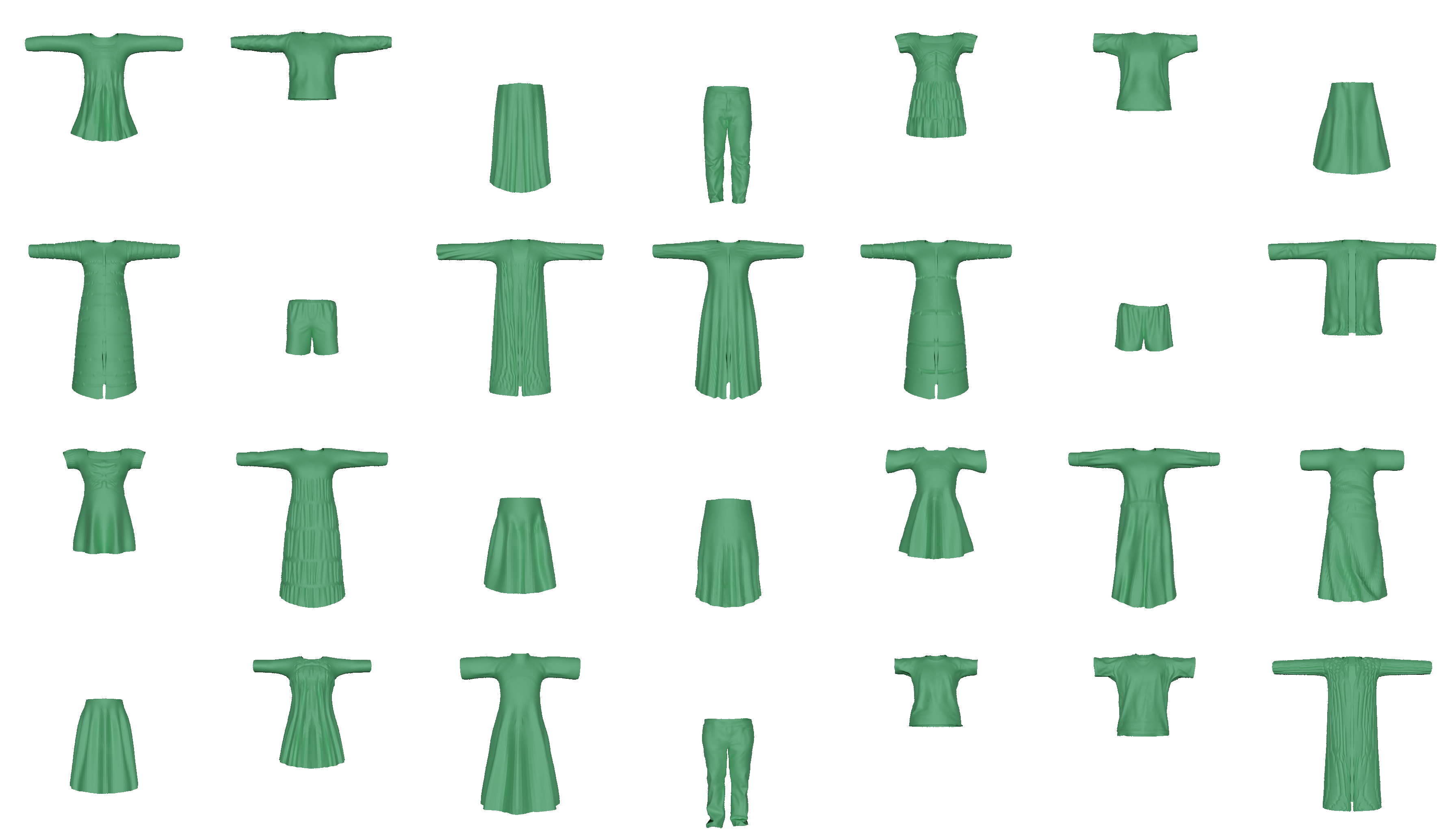}
  \caption{An illustration of our \textbf{Local Detail Database}.}
  \label{fig:LOD_1}
\end{figure*}
\begin{figure*}[htbp]
  \centering
  \includegraphics[width=.96\linewidth]{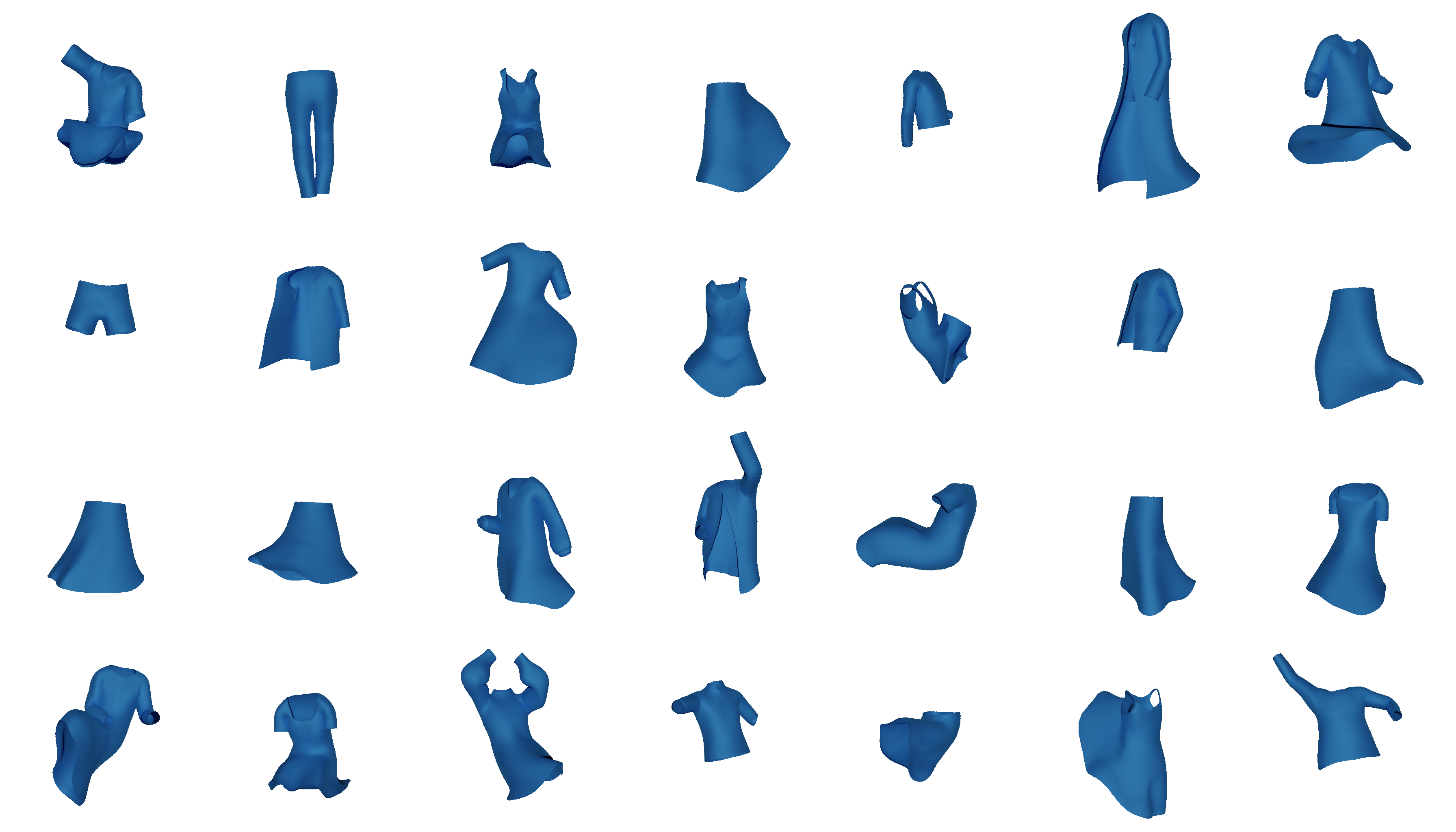}
  \caption{An illustration of our \textbf{Garment Deformation Database}.}
  \label{fig:LOD_2}
\end{figure*}
\begin{figure*}[htbp]
  \centering
  \includegraphics[width=.96\linewidth]{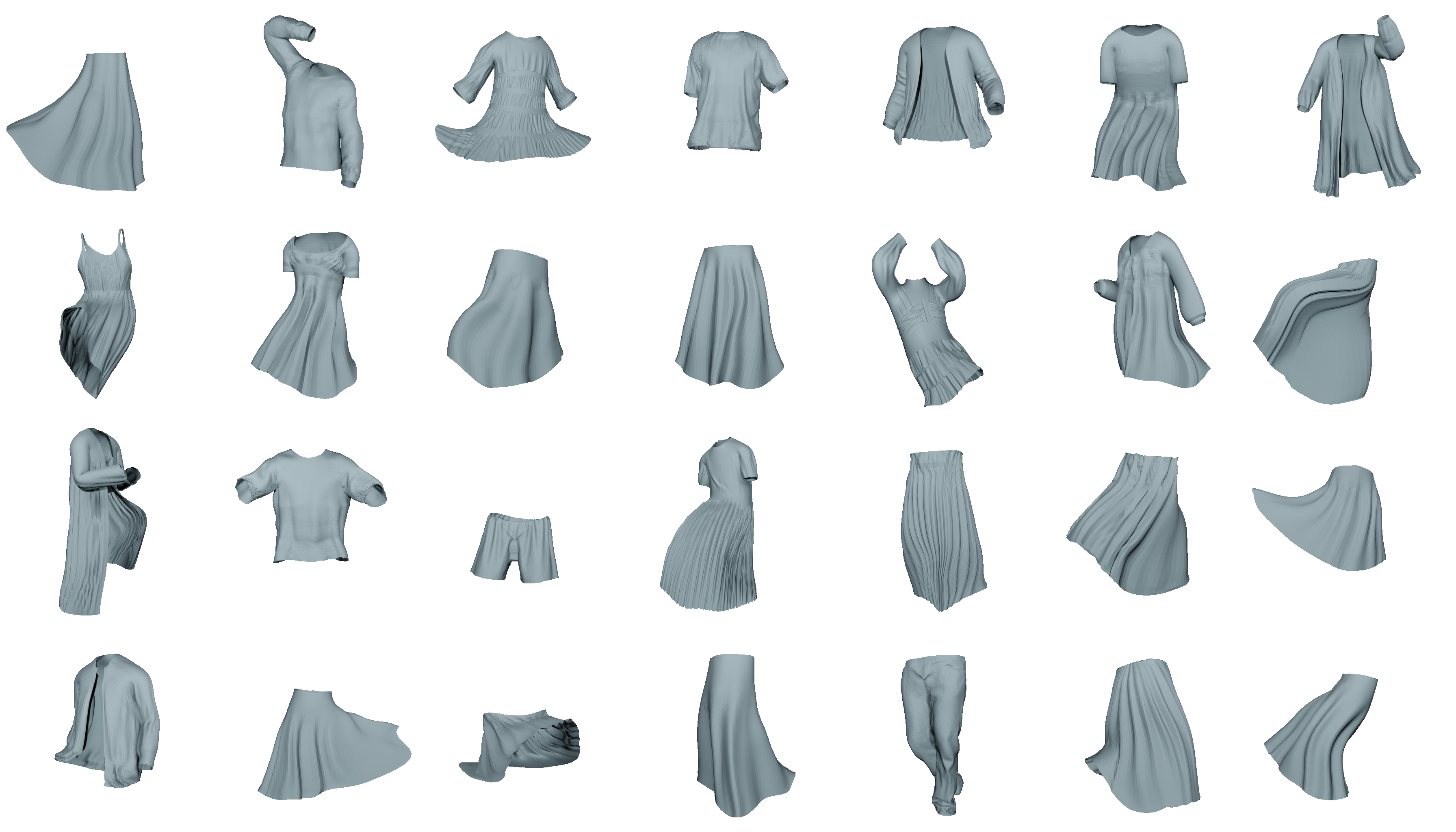}
  \caption{An illustration of our \textbf{Fine Garment Dataset}.}
  \label{fig:LOD_3}
\end{figure*}

%% file: supp_tables/table_notation.tex
\begin{table*}[htbp]
    \centering
    \caption{Explanation of notations used in the Main Paper.}
    \resizebox{0.56\textwidth}{!}{
        \begin{tabular}{lc}
        Notation & Description \\
        \midrule
        LOD & Levels of Details \\

        PCA & Principal Component Analysis  \\
        
        $M_C$ & Coarse garment sampled from the Garment Style Database \\

        $L_C$, $L_F$ & Garment pair that describes the local geometric detail \\

        $M_L$ & Garment after applying the local details from ($L_C$, $L_F$) to $M_C$ \\

        $D_T$, $D_F$ &  Garment pair that depicts global deformation \\

        $T$ & Deformation offsets of ($D_T$, $D_F$) in the rest-pose space \\
        
        LBS & Linear Blend Skinning \\ 

        $M_D$ &  Garment after transferring the deformation from ($D_T$, $D_F$) to $M_L$ \\

        $G(\cdot)$ & Statistical Garment Model worn on the mean shape of SMPL \\

        $\mathbf{T}_{\mathrm{g}}$ & Garment Template (i.e., The garment mean shape) \\

        $B_{\mathrm{g}}(\cdot)$ & Garment Shape Blend Shape (GSBS) in T-posed space \\

        $\alpha$ & The coefficients of $G(\cdot)$, which control the GSBS \\

        $T_{\mathrm{B}}(\cdot)$ & T-posed Body Mesh \\

        $\mathbf{T}_{\mathrm{b}}$ & Body Template (i.e., SMPL‘s mean shape) \\

        $B_{\mathrm{s}}(\cdot)$ & Body Shape Blend Shape (BSBS) of SMPL \\

        $B_{\mathrm{p}}(\cdot)$ & Body Pose Blend Shape (BPBS) of SMPL \\

        $\beta,\theta$ & The shape and pose parameters of SMPL \\

        $M_{\mathrm{B}}(\cdot)$ & Posed Body Mesh \\

        $W(\cdot)$ & Skinning Function \\

        $\mathcal{W}$ & Skinning Weights \\

        $J(\cdot)$ & Joint Locations \\

        $\widetilde{B}_{\mathrm{s}}(\cdot)$ & Garment displacements influenced by the BSBS, i.e., $B_{\mathrm{s}}(\cdot)$ \\

        $\widetilde{B}_{\mathrm{p}}(\cdot)$ & Garment displacements influenced by the BPBS, i.e., $B_{\mathrm{p}}(\cdot)$ \\

        $w(\cdot)$ & Weights for computing garment displacements and skinning \\

        $T_{\mathrm{G}}(\cdot)$ & T-posed garment after applying $\widetilde{B}_{\mathrm{s}}(\cdot)$ and $\widetilde{B}_{\mathrm{p}}(\cdot)$ to $G(\cdot)$ \\

        $\widetilde{\mathcal{W}}$ & Garment skinning weights extended from SMPL \\

        $M_P(\cdot)$ & Posed Garment Mesh \\

        $M_I$ & Fine garment predicted by the pixel-aligned network \\

        $p$ & Arbitrary point in 3D space\\ 
        
        $I_F(\cdot)$ & Pixel-aligned Features \\

        $\pi(\cdot)$ &  Projection Function\\
        
        $F(\cdot)$ & Feature Extraction Function \\

        $z(\cdot)$ & Depth value in the camera coordinate space \\

        $f(\cdot)$ & Implicit Function (MLP for decoding the occupancy of $p$) \\

        $s$ & The occupancy status of $p$ to the garment surface \\
        
        $\psi_{enc}$ & Triplane Encoder \\
        
        $\psi_{dec}$ & MLP-based decoder for decoding the occupancy of $p$ \\
        
        $G_F(\cdot)$ & Geometry-aware Features \\
        
        $F_{xy}, F_{xz}, F_{yz}$ & 3D axis-aligned features of three orthogonal planes \\
        
        $f_i(\cdot)$ & Implicit Function of the $i$-th boundary, i.e., $\psi_{dec}$ \\
        
        $o_i$ & The occupancy status of $p$ to the $i$-th boundary \\

        $L_{boundary}$ & Boundary Fitting Loss \\
        
        $L_{c}$ & Chamfer Distance Loss~\cite{ravi2020accelerating} \\
        
        $L_{lap}$ & Laplacian Smooth Regularization~\cite{ravi2020accelerating} \\

        $L_{edge}$ & Edge Length Regularization~\cite{ravi2020accelerating} \\

        $L_{normal}$ & Normal Consistency Regularization~\cite{ravi2020accelerating} \\
        
        $\lambda_{c}$, $\lambda_{lap}$, $\lambda_{edge}$, $\lambda_{normal}$ & Loss Weight \\

        $L_{nicp}$ & Registration Loss (i.e., loss for nicp) \\
        
        $L_{d}$ & Distance Cost: Deformed Shape vs. GT~\cite{amberg2007optimal} \\

        $L_{b}, L_{s}$ & Landmark Cost, Stiffness Term~\cite{amberg2007optimal} \\

        $L_{reg}$ & Mesh Regularization Terms \\

        $\lambda_{d}$, $\lambda_{b}$, $\lambda_{s}$, $\lambda_{reg}$ & Loss Weight
        
        \end{tabular}
    }
    \label{tab:notation}
\end{table*}